\documentclass[10pt,twocolumn,letterpaper]{article}

\usepackage{cvpr}              % To produce the CAMERA-READY version
\usepackage{multirow}
\usepackage{array}
\usepackage[dvipsnames]{xcolor}

\definecolor{cvprblue}{rgb}{0.21,0.49,0.74}
\usepackage[pagebackref,breaklinks,colorlinks,citecolor=cvprblue]{hyperref}

\def\paperID{13281} % *** Enter the Paper ID here
\def\confName{CVPR}
\def\confYear{2024}

\title{Boosting Image Restoration via Priors from Pre-trained Models}

\author{
	Xiaogang Xu$^{1,2,3}$ \quad Shu Kong$^{5,6,7}$ \quad Tao Hu$^{3,8}$ \quad Zhe Liu$^{1}$\footnotemark[1] \quad Hujun Bao$^{1,4}$ \\
	$^1$ Zhejiang Lab
	\quad $^2$ CUHK
	\quad $^3$ RealityEdge 
	\quad $^4$ Zhejiang University
	\quad $^5$ University of Macau \\
	$^6$  Institute of Collaborative Innovation 
	\quad $^7$ Texas A\&M University
	\quad $^8$ National University of Singapore\\
	{\tt \small xiaogangxu00@gmail.com, skong@um.edu.mo, yihouxiang@gmail.com} \\
	{\tt \small zhe.liu@zhejianglab.com, bao@cad.zju.edu.cn}
}

\begin{document}
\maketitle

\renewcommand{\thefootnote}{\fnsymbol{footnote}} 
\footnotetext[1]{Corresponding author.}

\begin{abstract}
Pre-trained models with large-scale training data, such as CLIP and Stable Diffusion, have demonstrated remarkable performance in various high-level computer vision tasks such as image understanding and generation from language descriptions. Yet, their potential for low-level tasks such as image restoration remains relatively unexplored.
In this paper, we explore such models to enhance image restoration.
As off-the-shelf features (OSF) from pre-trained models do not directly serve image restoration, we propose to learn an additional lightweight module called Pre-Train-Guided Refinement Module (PTG-RM) to refine restoration results of a target restoration network with OSF.
PTG-RM consists of two components, Pre-Train-Guided Spatial-Varying Enhancement (PTG-SVE), and Pre-Train-Guided Channel-Spatial Attention (PTG-CSA). PTG-SVE enables optimal short- and long-range neural operations, while PTG-CSA enhances spatial-channel attention for restoration-related learning.
Extensive experiments demonstrate that PTG-RM, with its compact size ($<$1M parameters), effectively enhances restoration performance of various models across different tasks, including low-light enhancement, deraining, deblurring, and denoising.
\end{abstract}

\section{Introduction}
\label{sec:intro}

%{\bf Background}.
Image restoration plays a vital role in real-world scenarios, aiming to reconstruct high-quality images by eliminating degradations. It has broad applications in various fields, such as denoising~\cite{xu2022pvdd,xu2022general} and low-light enhancement~\cite{xu2022snr,wu2023learning} for improving smartphone-captured photos.
While effective restoration networks have been proposed~\cite{li2023efficient,zamir2022restormer}, the inherently ill-posed nature of image restoration makes it challenging to achieve significant improvements by merely modifying network structures. Simply increasing model parameters does not guarantee better results, as the model may tend to overfit to the training data.

\begin{figure}[t]
	\begin{center}
		\includegraphics[width=1.0\linewidth]{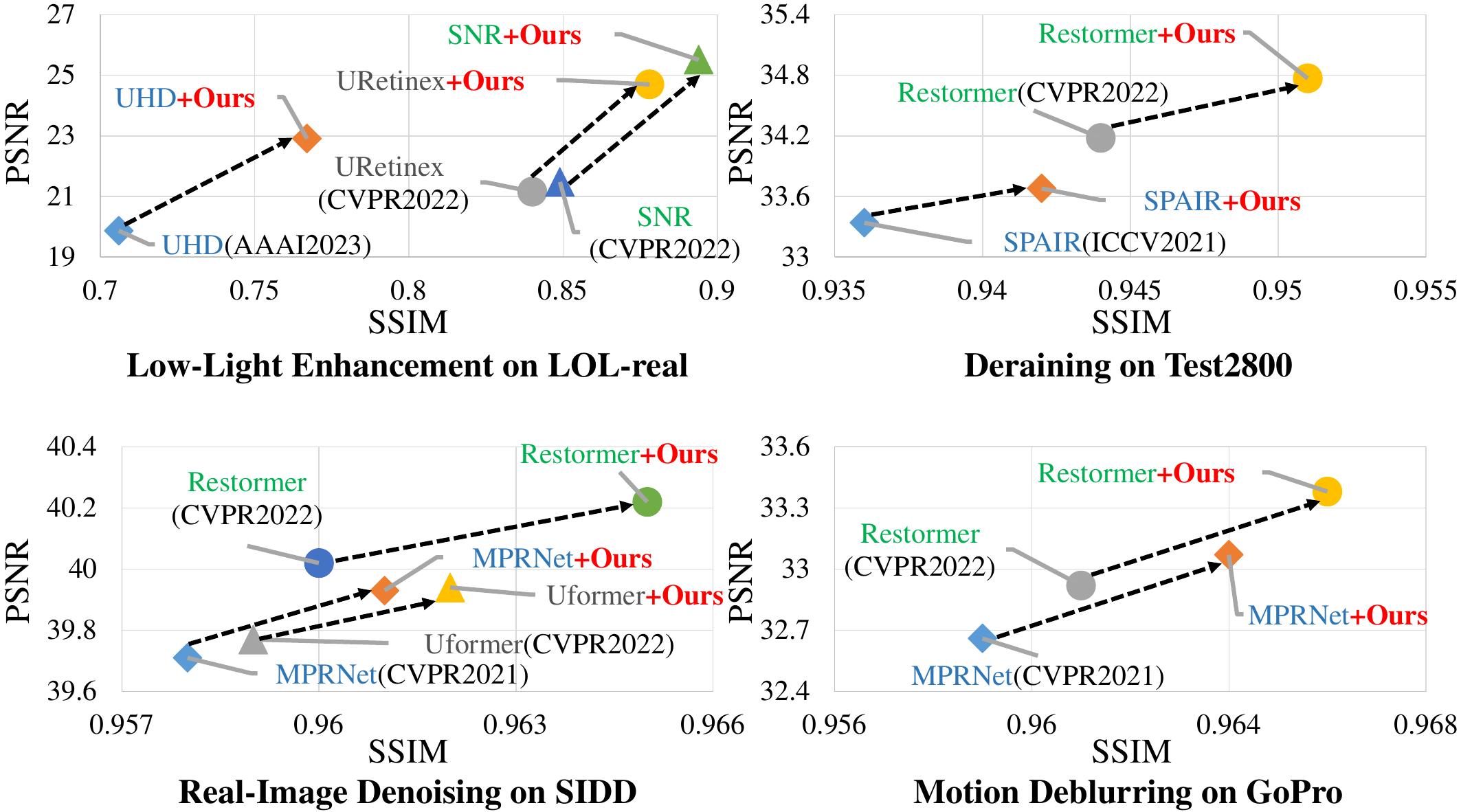}
\vspace{-0.35in}
 \end{center}
\caption{
 Our method leverages pre-trained models, such as CLIP and Stable Diffusion, and significantly improves image restoration across various tasks. More results on different tasks/models can be seen in experiments. Pre-trained models are involved during the training and not required during the inference.
}
	\label{fig:teaser}
 \vspace{-0.1in}
\end{figure}

\begin{figure*}[t]
\begin{center}
		\includegraphics[width=1.0\linewidth]{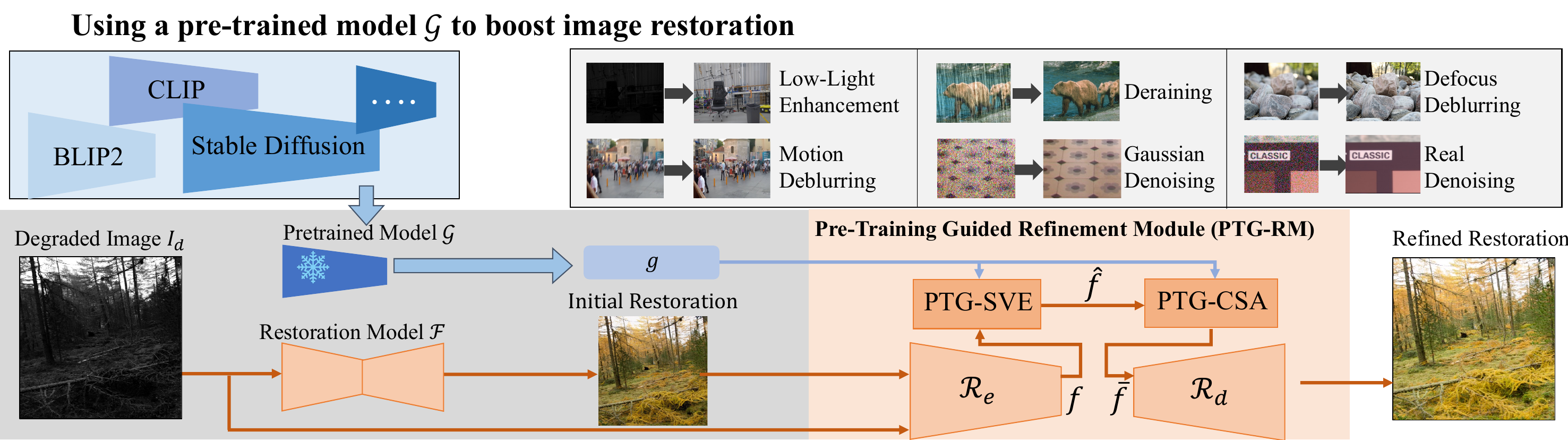}
	\end{center}
\vspace{-0.1in}
\caption{
We present a lightweight plugin, \emph{pre-training guided refining module} (PTG-RM), to leverage pre-trained models for enhancing image restoration. 
The desired prior is the OFS $\mathcal{G}(I_d)$.
It has two components, PTG spatial varying enhancement (PTG-SVE), and PTG channel-spatial attention (PTG-CSA). Fig.~\ref{fig:framework} depicts their details.
Our PTG-RM significantly improves restoration in various tasks as listed in the top-right (see quantitative results previewed in Fig.~\ref{fig:teaser}).
	}
\label{fig:teaser1}
\end{figure*}

Restoration performance relies on  strong image priors, such as the novel level of denoising~\cite{wang2022blind2unblind} or the blur kernel in deblurring~\cite{kong2018image, yang2023k3dn}. However, estimating these priors is challenging, especially with real-world data. Some approaches utilize physical variables as priors, like depth information~\cite{xu2023low} and semantic features~\cite{wu2023learning,aakerberg2022semantic,wang2018recovering} derived from pre-trained networks.
Nevertheless, these physical variables are not robust enough since the dense depth/semantic prediction networks do not have sufficient generalization ability among different scenes in restoration tasks.
As a result, employing them requires complex and specific mechanisms, limiting their applicability across various tasks. In this paper, we propose a novel approach that extracts degradation-related information from pre-trained models (with various training objectives) exposed to different degradation during pre-training, all without requiring explicit annotations.

{\bf Motivation}.
Two types of pre-trained models may contain degradation-related information during training: restoration models, and pre-trained models on large-scale data (e.g.,  CLIP~\cite{radford2021learning}, BLIP~\cite{li2022blip}, and BLIP2~\cite{li2023blip}). 
Using the former is evident, but models trained with some types of degradation may not effectively help restore images with other types of degradation.
Using the latter remains unexplored.  CLIP-IQA~\cite{wang2022exploring} finds that CLIP features contain degradation-related information and so be useful for image assessment, while no restoration approaches have been proposed yet.
Existing pre-trained multi-modality models may have been trained on various degraded images. 
Presumably, restoration-related annotations are unavailable during pre-training, their resulted features likely contain valuable information for image restoration.
The key is to leverage such information to help the target restoration learning.
However, the heterogeneity of pre-trained models and restoration models poses difficulties in using the off-the-shelf features extracted from pre-trained models.

{\bf Technical novelty}.
We introduce a novel pre-training guided refinement module (PTG-RM) that leverages off-the-shelf features (OSF) computed by a pre-trained model $\mathcal{G}$ to improve image restoration tasks.
The PTG-RM $\mathcal{R}$ is a  lightweight plugin (Fig.~\ref{fig:teaser1}) (additional $\mathcal{R}$ has $<$1M parameters in total).
PTG-RM enables us to determine optimal operation ranges and spatial-channel attention, thus facilitating image restoration.
It takes as input the initially enhanced image from $\mathcal{F}$, the input image, and its OSF extracted by a pre-trained model.
It is trained with $\mathcal{F}$ (using the same loss as $\mathcal{F}$) and adaptively enhances it.
PTG-RM $\mathcal{R}$ consists of two components: Pre-Train-Guided Spatial Varying Enhancement (PTG-SVE), and Pre-Train-Guided Channel-Spatial Attention (PTG-CSA).

PTG-SVE employs spatial-varying operations to refine the initially enhanced results differently from region to region.
Unlike previous methods~\cite{xu2022snr} that rely on fixed references to determine optimal operation ranges, we establish a spatial-aware learnable mapping for OSF and utilize the mapped features as spatial-wise guidance. This adaptively fuses the features extracted from short- and long-range operations, allowing different regions to be refined appropriatel and yielding more effective enhancement.

Following PTG-SVE, PTG-CSA further enhances the results by formulating effective channel- and spatial-attention with OSF.
We note that different areas may require varying degrees of feature correctness via the attention mechanism. Hence, we propose to generate  spatial-varying convolution kernels to synthesize the spatial weights. Our approach tailors the attention process to different regions.

{\bf Contributions}. We make three major contributions.
\begin{itemize}
    \item We present a novel and general method that leverages pre-trained models to enhance various restoration tasks. 
    Our work opens up possibilities for improving performance across various domains.
    \item We propose a novel paradigm that leverages pre-trained priors to formulate effective neural operation ranges and attention mechanisms. 
    \item We validate our method through extensive experiments on different datasets, networks, and tasks, and show remarkable improvements over prior methods (cf. Fig.~\ref{fig:teaser}).
\end{itemize}

\section{Related Work}

\textbf{Image Priors for Restoration.}
Different restoration tasks demand distinct image priors, such as noise levels for denoising  and blurring kernels for deblurring. 
Due to the ill-posed nature of restoration, estimating priors is difficult. In real-world scenarios, these priors are typically intertwined, adding further complexity to the restoration process.
Recent literature introduces several methods to improve restoration by leveraging multi-modal maps as unified priors. These methods predominantly rely on pre-computed physical multi-modal maps. For instance, SKF~\cite{wu2023learning} uses semantic maps to optimize the feature space for low-light enhancement. SMG~\cite{xu2023low} employs a generative framework to integrate edge, depth, and semantic information, enhancing the initial appearance modeling for low-light scenarios. Additionally, some approaches use Near-Infrared (NIR) information to refine imaging results~\cite{jin2022darkvisionnet,wan2022purifying}. 
These priors are also applied to other restoration tasks, such as image denoising~\cite{liu2017image} and deraining~\cite{li2022close}.
However, aligning these priors with the input image can be challenging, and errors in the priors may adversely impact performance.
Different from existing works, we propose to leverage pre-trained models as priors to enhance image restoration.

\vspace{1mm}
\textbf{Pre-Trained Models for Downstream Tasks.}
Recently, a series of pre-trained models with large-scale training datasets have emerged, particularly in the form of multi-modal models such as CLIP~\cite{radford2021learning}, BLIP~\cite{li2022blip}, and BLIP2~\cite{li2023blip}.
The feature space learned by these models offers rich knowledge that can benefit various tasks. While previous work has demonstrated the effectiveness of CLIP in high-level tasks like zero-shot classification~\cite{esmaeilpour2022zero,zhai2022lit}, image editing~\cite{patashnik2021styleclip,avrahami2022blended}, open-world segmentation~\cite{zhou2023zegclip,wang2022cris}, and 3D classification~\cite{zhang2022pointclip,xue2022ulip}, its potential for aiding low-level restoration tasks remains unexplored. Only the capability of employing such for image quality assessment, as demonstrated in CLIP-IQA, has been explored.
We propose a general framework to leverage pre-trained models to improve various restoration tasks.

\section{Methods}

\textbf{Background.} Let $I_d$ represent a degraded image, and $I_c$ denote the corresponding ground-truth (without degradation). 
A restoration network $\mathcal{F}$ produces restored image $\hat{I}_c=\mathcal{F}(I_d)$. 
Despite the existence of various effective network structures $\mathcal{F}$ that have been proposed, there are current upper bounds in these tasks. 
Breaking through these bounds often requires designing more complex networks and training strategies, which can be arduous.
Additionally, innovations in network architecture or training strategies for one task might not translate to another.
While different priors $g$ have been introduced into the restoration process, including image and physical priors, estimating these priors is difficult.

\textbf{Motivation.} 
We hypothesize that the prior $g$ can be effectively represented as the feature extracted from various pre-trained models $\mathcal{G}$, as $g=\mathcal{G}(I_d)$.
Note that $\mathcal{G}$ is typically not trained with restoration targets but might have been exposed to images with diverse degradations.
So it is likely to learn useful information to help image restoration.
We propose a novel approach that uses $g$ to improve the initial restoration by $\mathcal{F}$, even if these networks have already reached their current upper bounds.

\textbf{Challenge.} 
Using $g$ to assist $\mathcal{F}$ is non-trivial. Primarily, the feature $g$ is not inherently aligned with the restoration tasks because they might represent different aspects. For instance, features from CLIP focus more on semantic information, making direct alignment to restoration challenging. 
Moreover, these priors exhibit varying shapes, such as the one-dimensional (1D) features from the CLIP model, while the features in $\mathcal{F}$ are typically 2D. To reconcile the discrepancies in both representation and shape, we propose a refinement module $\mathcal{R}$ to refine the initial restoration by $\mathcal{F}$. 
This eliminates the need to align $g$ to distinct features of $\mathcal{F}$ and allows for a unified 1D representation for $g$.
Furthermore, we introduce a novel approach to utilize $g$ to formulate optimal neural operating ranges via an effective attention mechanism in $\mathcal{R}$. 
This implicitly distills restoration-related information, effectively boosting the final performance.

\begin{figure*}[t]
	\begin{center}
  \includegraphics[width=1.0\linewidth]{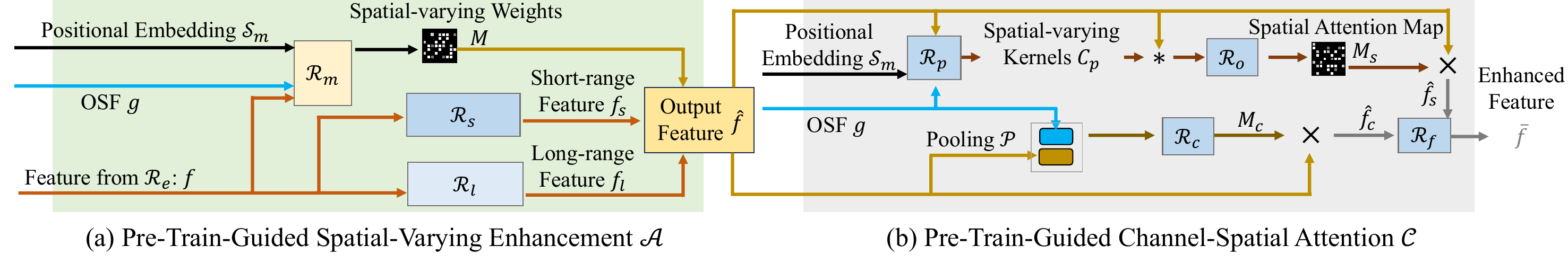}
	\end{center}
	\vspace{-0.2in}
	\caption{The pipeline of PTG-SVE and PTG-CSA. In PTG-SVE, we use the learnable spatial embedding $\mathcal{S}_m$, OSF $g$, and input feature $f$ to adaptively formulate spatial weights ($M$, Eq.~\ref{eq:mra}) for fusing short- and long-range processed features ($f_s$ and $f_l$) via operations $\mathcal{R}_s$ and $\mathcal{R}_l$, yielding $\hat{f}$ (Eq.~\ref{eq:fsl}). In PTG-CSA, OSF $g$ conditions channel attention $M_c$ for $\hat{f}$ through $\mathcal{R}_c$ (Eq.~\ref{eq:mc}). Additionally, $g$ combines with learnable spatial representation $\mathcal{S}_c$ and $\hat{f}$ to generate spatial attention map $M_s$, using spatial-wise convolutions $C_p$ (obtained via $\mathcal{R}_p$) to derive $\hat{\mathcal{M}}_s$ that is further processed with $\mathcal{R}_o$ (Eqs.~\ref{eq:cp} and~\ref{eq:ms}). Channel- and spatial-attention outputs ($\hat{f}_c$ and $\hat{f}_s$) merge via $\mathcal{R}_f$ to enhance feature $\bar{f}$ (Eq.~\ref{eq:barf}).
	}
	\label{fig:framework}
\end{figure*}

\subsection{Overview of Refinement Module}
Fig.~\ref{fig:teaser1} depicts the restoration pipeline using our method.
Given an input image $I_d$, we have an initial restoration result as $\hat{I}_c=\mathcal{F}(I_d)$. 
We aim to refine the result using the proposed pre-training guided refinement module (PTG-RM) $\mathcal{R}$, resulting in $\bar{I}_c=\mathcal{R}(\hat{I}_c, I_d, g)$. The key of this approach is to distill restoration-related information from the prior $g$.

$\mathcal{R}$ is a simple encoder-decoder structure. The encoder and decoder of $\mathcal{R}$ are denoted as $\mathcal{R}_e$ and $\mathcal{R}_d$, respectively. To ensure lightweight implementation, distillation occurs in the latent space, avoiding the need to align $g$ with restoration-related features.
The latent feature $f$ is derived through a comparison between the initial enhanced results and the original input images, given as $f=\mathcal{R}_e(\hat{I}_c \oplus I_d)$, where $\oplus$ denotes the concatenation operation. The resulting $f$ is in $\mathbb{R}^{h\times w \times c}$, with $h$, $w$, and $c$ representing feature height, width, and channel number, respectively.
The priors are used in further learning the latent feature as $\bar{f}=\mathcal{C}(\mathcal{A}(f, g), g)$, where $\mathcal{A}$ and $\mathcal{C}$ represent the Pre-Train-Guided Spatial-Varying Enhancement (PTG-SVE) and Pre-Train-Guided Channel-Spatial Attention (PTG-CSA) modules, respectively.
The final enhancement is obtained from the decoder as $[I_m, I_r]=\mathcal{R}_d(\hat{f})$, comprising two components. The first component, $I_m$, represents the correction mask used to mitigate errors in the initial enhancement results. The second component, $I_r$, is the residual refinement that addresses artifacts and adds additional details.
The final result is denoted as 
\begin{equation}
    \bar{I}_c=I_d + (\hat{I}_c-I_d) \times I_m + I_r.
    \label{eq:refinement}
\end{equation}

\subsection{Pre-Train-Guided Spatial-Varying Operations}

In PTG-SVE, we argue that $g=\mathcal{G}(I_d)$ may contain information reflecting the pixel-level image quality of $I_d$. For areas with poor quality, long-range operations are used to capture non-local features, while regions with relatively good quality prioritize local features for accurate restoration.

In Fig.~\ref{fig:framework}, the primary objective is to predict the optimal neural operation range for each location of the feature map $f$, which we refer to as the ``range score map'', denoted as $M$. To ensure a general $\mathcal{R}$ with unified 1D priors $g$ from various models, we propose adding location-aware embeddings for the priors, 
thereby adaptively discovering quality information for different pixels. Let $S=\{(x,y)|x\in[1, w], y\in [1,h]\}$ represent the 2D coordinate map with dimensions $h\times w\times 2$.
We use a position embedding module $\mathcal{P}$ to generate spatial representation, denoted as $\mathcal{S}_m=\mathcal{P}(S)$, where $\mathcal{S}\in \mathbb{R}^{h\times w \times c}$.
Furthermore, to determine the admired neural operation range for each location of $f$, we use a learnable mapping function $\mathcal{T}_m$ to transform the priors to another space that can more effectively decide the optimal range.
To obtain $M$, we use a range-learning module $\mathcal{R}_m$, which takes the encoder's feature $f$, the pre-trained prior $g$, and the spatial representation $\mathcal{S}_m$ as inputs.
The procedure is denoted as 
\begin{equation}
    M=\mathcal{R}_m(f\oplus \mathcal{T}_m(g) \oplus \mathcal{S}_m).
    \label{eq:mra}
\end{equation}

Following~\cite{xu2022snr}, we use CNN for the short-range operation, denoted as $\mathcal{R}_s$, and transformer for the long-range operation, represented as $\mathcal{R}_l$. Specifically, we employ the Restormer backbone for $\mathcal{R}_l$ and ResNet for $\mathcal{R}_s$. 
Suppose the features after the short- and long-range operation are $f_s$ and $f_l$, respectively. We can obtain the refined feature $\hat{f}$ as 
\begin{equation}
f_s=\mathcal{R}_s(f), \, f_l=\mathcal{R}_l(f), \, \hat{f}=M\times f_s+(1-M)\times f_l.
\label{eq:fsl}
\end{equation}
The previous approach \cite{xu2022snr} relies on pre-computed SNR values, which may not always be accurate and can fail to enhance results, especially when the initial results from $\mathcal{F}$ have reached their upper bound.
In contrast, our score range map is learned online based on the input image, restoration-related priors, and explicit spatial features that are learnable. This flexibility allows us to handle various situations, resulting in better performance and generalization (as demonstrated in the ablation study).

\begin{table*}[t]
	\centering
\Large
\resizebox{1.0\linewidth}{!}{
		\begin{tabular}{cc|cc|cc|cc|cc|cc|cc}
			\toprule[1pt]
\multirow{2}{1.2cm}{\textbf{Datasets}} &  \multirow{2}{1.2cm}{\textbf{Methods}} & \multicolumn{2}{c|}{\textbf{Original}} & \multicolumn{2}{c|}{\textbf{+Ours-c}} & \multicolumn{2}{c|}{\textbf{+Ours-b}} & \multicolumn{2}{c|}{\textbf{+Ours-s}} & \multicolumn{2}{c|}{\textbf{+Ours-r}} & \multicolumn{2}{c}{\textbf{+Ours-f}}\\

& & PSNR & SSIM& PSNR & SSIM& PSNR & SSIM& PSNR & SSIM& PSNR & SSIM& PSNR & SSIM\\ \hline
\multirow{3}{1cm}{\textbf{LOL}} & UHD & 19.87 & 0.706 & 22.91 (\textbf{+3.04}) & 0.767 (\textbf{+6.1}) & 21.83 (\textbf{+1.96}) & 0.732 (\textbf{+2.6}) & 22.35 (\textbf{+2.48}) & 0.758 (\textbf{+5.2}) & 21.71 (\textbf{+1.84}) & 0.737 (\textbf{+3.1})&22.74 (\textbf{+2.87})&0.764 (\textbf{+5.8}) \\
 & URetinex & 21.16 & 0.840 & 24.70 (\textbf{+3.54})&0.878 (\textbf{+3.8})&23.57 (\textbf{+2.41})&0.869 (\textbf{+2.9})&24.23 (\textbf{+3.07})&0.866 (\textbf{+2.6})&23.99 (\textbf{+2.83})&0.862 (\textbf{+2.2})&24.56 (\textbf{+3.40})&0.870 (\textbf{+3.0})\\
 & SNR & 21.48 & 0.849 &25.50 (\textbf{+4.02})&0.892 (\textbf{+4.3})&25.61 (\textbf{+4.13})&0.891 (\textbf{+4.2})&25.19 (\textbf{+3.71})&0.874 (\textbf{+2.5})&25.24 (\textbf{+3.76})&0.887 (\textbf{+3.8})&24.90 (\textbf{+3.42})&0.888 (\textbf{+3.9})\\ \hline
 \multirow{3}{1cm}{\textbf{SID}} & UHD &20.46&0.614 &20.99 (\textbf{+0.53})&0.616 (\textbf{+0.2})&21.06 (\textbf{+0.60})&0.619 (\textbf{+0.5})&22.34 (\textbf{+1.88})&0.625 (\textbf{+1.1})&21.11 (\textbf{+0.65})&0.618 (\textbf{+0.4})&21.08 (\textbf{+0.62})&0.619 (\textbf{+0.5})\\
 & URetinex & 21.56 & 0.619 & 22.34 (\textbf{+0.78}) &0.623 (\textbf{+0.4})&22.02 (\textbf{+0.46})&0.621 (\textbf{+0.2})&22.21 (\textbf{+0.65})&0.623 (\textbf{+0.4})&22.17 (\textbf{+0.61})&0.625 (\textbf{+0.6})&22.40 (\textbf{+0.84})&0.626 (\textbf{+0.7})\\
&SNR&22.87&0.625&23.34 (\textbf{+0.47})&0.630 (\textbf{+0.5}) &23.15 (\textbf{+0.28})&0.627 (\textbf{+0.2})&23.08 (\textbf{+0.21})&0.631 (\textbf{+0.6})&23.06 (\textbf{+0.19})&0.632 (\textbf{+0.7}) &23.17 (\textbf{+0.30})&0.636 (\textbf{+1.1})\\
			\bottomrule[1pt]
	\end{tabular}}
\vspace{-0.1in}
	\caption{Comparisons on LOL-real and SID dataset. $-c$, $-b$, $-s$, and $-r$ refer to using CLIP, BLIP2, Stable Diffusion, and restoration models trained on SDSD, respectively. $-f$ denotes applying refinement on the features of $\mathcal{F}$. (+) indicates improvements for PSNR and SSIM$_{({\rm x}100)}$.}
	\label{comparison1}
\end{table*}

\subsection{Pre-Train-Guided Attention}
As shown in Fig.~\ref{fig:framework}, we further introduce a lightweight component that utilizes pre-trained priors $g$ to create an effective attention mechanism in $\mathcal{R}$. Optimizing the feature attention in $\mathcal{R}$ is crucial for effectively identifying helpful features to enhance the initial results $\hat{I}_c$. This involves both spatial-level and channel-level attentions.
The hidden restoration-related information in $g$ can be discovered by using $g$ to improve the restoration features in $\mathcal{R}$ conditioned on them.

We begin by formulating the attention computation at the channel level. We introduce a mapping function $\mathcal{T}_c$ to transform $g$ into the attention-prediction space, and utilize the channel attention computation module $\mathcal{R}_c$.
The formulation of the channel attention is 
\begin{equation}
    \begin{aligned}
        &M_c=\mathcal{R}_c(\mathcal{O}(\hat{f}) \oplus \mathcal{T}_c(g)), \; \hat{f}_c=\hat{f} \times M_c,
    \end{aligned}
    \label{eq:mc}
\end{equation}
where $\mathcal{O}$ is the pooling operation, and $\mathcal{M}_c \in \mathbb{R}^{c}$.

As for the spatial-attention computation, we utilize the 1D pre-trained prior $g$ to predict location-wise attention based on the feature distribution of each location in $\hat{f}$. Simply using the spatial location information, as shown in Eq.~\ref{eq:mra}, results in each pixel's feature considering a similar condition for neighboring features, limiting the elimination of spatial artifacts. 
Therefore, we propose an alternative strategy by predicting the neural operation parameters for each location, optimizing the spatial attention based on the varying location-wise feature distribution. We denote the spatial attention computation module as $\mathcal{R}_p$, and first formulate the location-wise convolution map, as
\begin{equation}
    \mathcal{C}_p=\mathcal{R}_p(\hat{f}, \mathcal{T}_c(g), \mathcal{S}_c),
\label{eq:cp}
\end{equation}
where the obtained convolution map $\mathcal{C}_p\in \mathbb{R}^{h\times w\times (k_h\times k_w\times c)}$, $k_h$ and $k_w$ are the convolution kernel size, and $\mathcal{S}_c$ is another learnable position embedding here.
The obtained convolution maps can be utilized to optimize the feature, and spatial attention can be obtained as
\begin{equation}
\hat{M}_s=\hat{f} * \mathcal{C}_p, \, M_s=\mathcal{R}_o(\hat{M}_s),
\label{eq:ms}
\end{equation}
where $*$ is the convolution operation for each location, and $\mathcal{R}_o$ is another learnable operation which mapps the feature channel $c$ to 1, eliminating the influence from the channel-level dependency.
Further, the feature after spatial attention can be described as $\hat{f}_s=\hat{f}\times M_s$.

The features after spatial and channel attentions can be merged via a fusion module as
\begin{equation}
    \bar{f}=\mathcal{R}_f(\hat{f}_c \oplus \hat{f}_s),
    \label{eq:barf}
\end{equation}
where $\mathcal{R}_f$ denotes the fusin module. 
The obtained feature $\bar{f}$ can be processed via a decoder $\mathcal{R}_d$ to obtain the residual refinement $I_r$ and the mask $I_m$ as indicated in Eq.~\ref{eq:refinement}.

\subsection{Loss Function}
Our designed $\mathcal{R}$ can be jointly trained with the model $\mathcal{F}$. Suppose the paired ground truth for the input image $I_d$ is $\mathcal{I}_c$, and the loss function for the model $\mathcal{F}$ is denoted as $\mathcal{L}_g(\hat{I}_c, \mathcal{I}_c)$ (is usually the reconstruction loss in the pixel level or perceptual loss, and can also be the unsupervised loss), then the loss function for the refinement module can be written as $\mathcal{L}_g(\bar{I}_c, \mathcal{I}_c)$. 
In summary, the overall loss is 
\begin{equation}
    \mathcal{L}_g(\hat{I}_c, \mathcal{I}_c)+\lambda_1 \mathcal{L}_g(\bar{I}_c, \mathcal{I}_c),
\end{equation}
where $\lambda_1$ is the loss weight and remains robust across various tasks and networks (in our experiments, $\lambda1$ is always set as 1).

\begin{table}[t]
	\centering
	\small
		\resizebox{1.0\linewidth}{!}{
		\begin{tabular}{c|p{1.5cm}<{\centering}p{1.5cm}<{\centering}p{1.5cm}<{\centering}p{1.5cm}<{\centering}p{1.5cm}<{\centering}}
			\toprule[1pt]
			Methods & SNR & +SKF & +SMG & +SMG(dep) &+Ours-c\\
                \hline
			PSNR  &21.48 & 23.05 & 24.84& 24.12&\textbf{25.50}\\
			SSIM & 0.849 &0.853 &0.880 &0.851 &\textbf{0.892}\\
    \hline
   Methods & URetinex & +SKF & +SMG & +SMG(dep) &+Ours-c\\
   PSNR &21.16 &23.51 &23.74 &23.25 &\textbf{24.70} \\
   SSIM & 0.840&0.856 &0.852 &0.849 &\textbf{0.878} \\
   \hline
   +Params & 0 & 2.15M & 16.76M & 16.76M & 0.67M\\ 
			\bottomrule[1pt]
	\end{tabular}}
\vspace{-0.1in}
	\caption{Quantitative comparison on the LOL-real dataset. +Params means the additional parameter number compared with original $\mathcal{F}$.}
  \vspace{-0.1in}
	\label{comparison3}
\end{table}

\section{Experiments}
We first introduce tasks and datasets used in experiments, followed by a detailed analysis of our methd using low-light image enhancement as an example. We also demonstrate the effectiveness of our method on other tasks.

\subsection{Tasks and Datasets}
For low-light enhancement, we use the SID~\cite{chen2018learning} and LOL-real~\cite{yang2021sparse} datasets.
For deraining, we use the Rain13K~\cite{zamir2022restormer} dataset for training and test on Rain100H~\cite{yang2017deep}, Rain100L~\cite{yang2017deep}, Test100~\cite{zhang2019image}, Test1200~\cite{zhang2018density}, and Test2800~\cite{fu2017removing} datasets.
For gaussian denoising, we use two settings: synthetic noise on Set12~\cite{DnCNN}, BSD68~\cite{martin2001database_bsd}, CBSD68~\cite{martin2001database_bsd}, Kodak~\cite{kodak}, McMaster~\cite{zhang2011color_mcmaster}, and Urban100~\cite{huang2015single_urban100}; and real-world denoising on SIDD~\cite{sidd}.
For single-image motion deblurring, we use the GoPro~\cite{gopro2017} dataset for training and evaluate on synthetic datasets (GoPro~\cite{gopro2017}, HIDE~\cite{shen2019human}) and real-world datasets (RealBlur-R~\cite{rim_2020_realblur}, RealBlur-J~\cite{rim_2020_realblur}).
For defocus deblurring, we use the DPDD~\cite{abdullah2020dpdd} training data and test on the EBDB~\cite{karaali2017edge_EBDB} and JNB~\cite{shi2015just_jnb} datasets.

\begin{figure}[t]
	\centering
 \begin{subfigure}[c]{0.11\textwidth}
		\centering
		\includegraphics[width=0.8in]{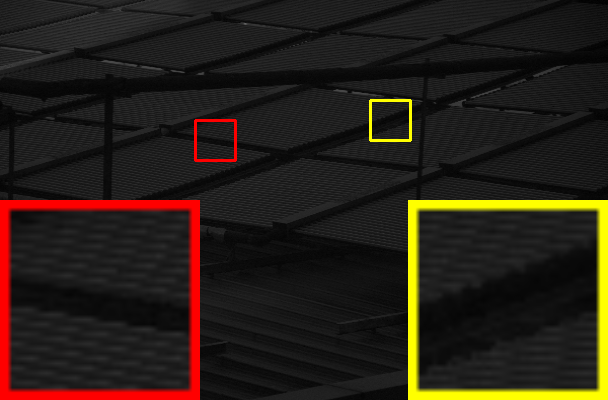}
		\vspace{-1.5em}
		\caption*{\small Input}
	\end{subfigure}
	\begin{subfigure}[c]{0.11\textwidth}
		\centering
		\includegraphics[width=0.8in]{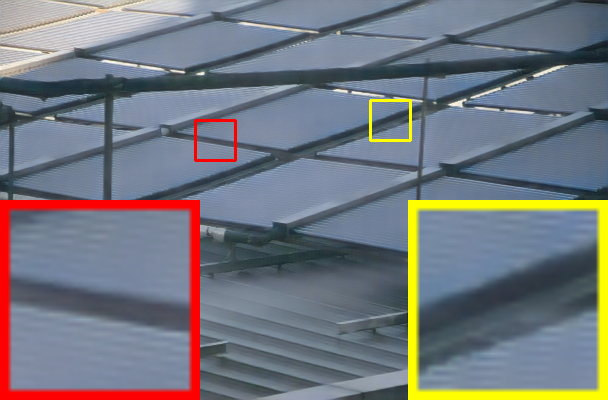}
		\vspace{-1.5em}
		\caption*{\small SNR}
	\end{subfigure}
	\begin{subfigure}[c]{0.11\textwidth}
		\centering
		\includegraphics[width=0.8in]{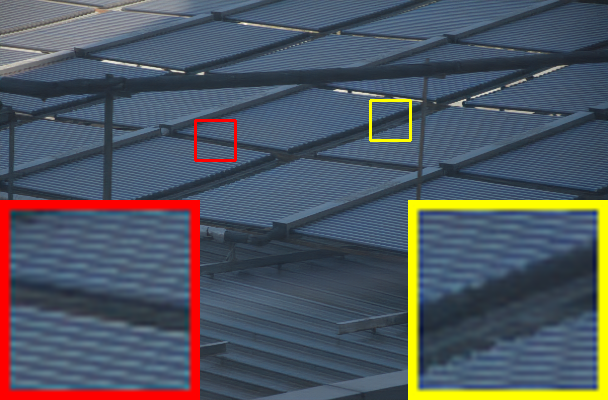}
		\vspace{-1.5em}
		\caption*{\small  \bf SNR+Ours}
	\end{subfigure}
	\begin{subfigure}[c]{0.11\textwidth}
		\centering
		\includegraphics[width=0.8in]{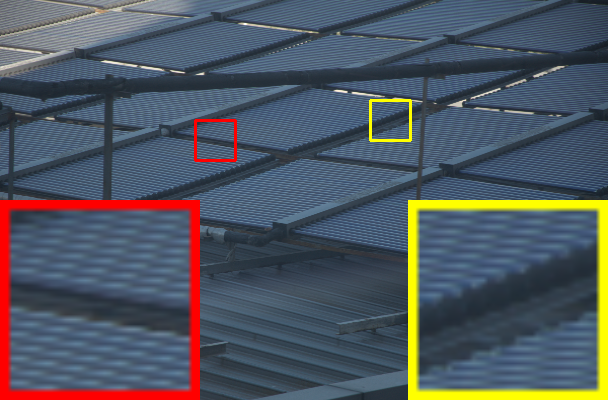}
		\vspace{-1.5em}
		\caption*{Ground-truth}
	\end{subfigure}  
	\vspace{0.2em} \\

 \begin{subfigure}[c]{0.11\textwidth}
		\centering
		\includegraphics[width=0.8in]{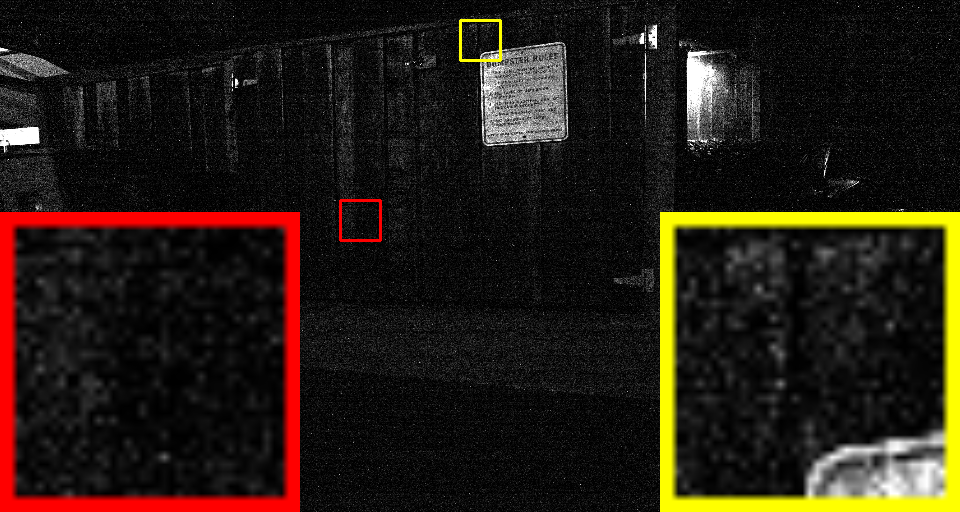}
		\vspace{-1.5em}
		\caption*{\small Input}
	\end{subfigure}
	\begin{subfigure}[c]{0.11\textwidth}
		\centering
		\includegraphics[width=0.8in]{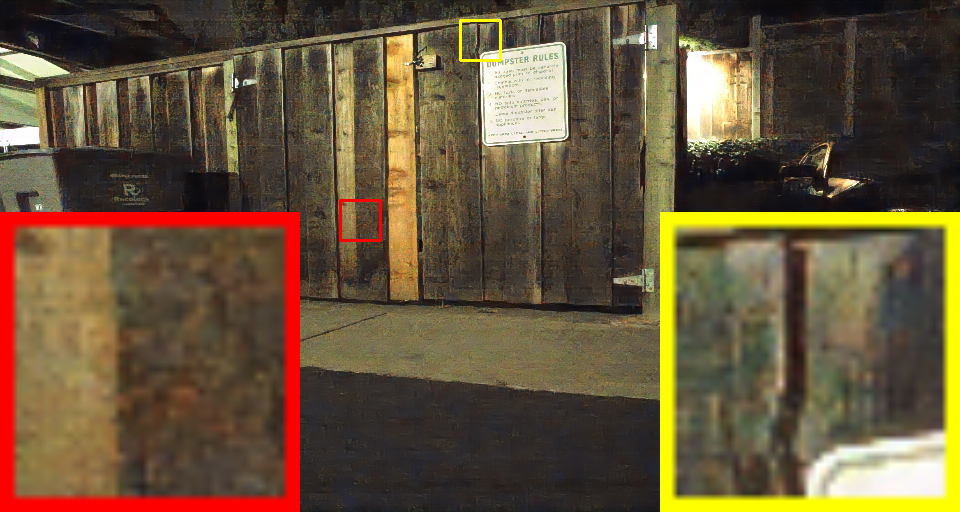}
		\vspace{-1.5em}
		\caption*{\small SNR}
	\end{subfigure}
	\begin{subfigure}[c]{0.11\textwidth}
		\centering
		\includegraphics[width=0.8in]{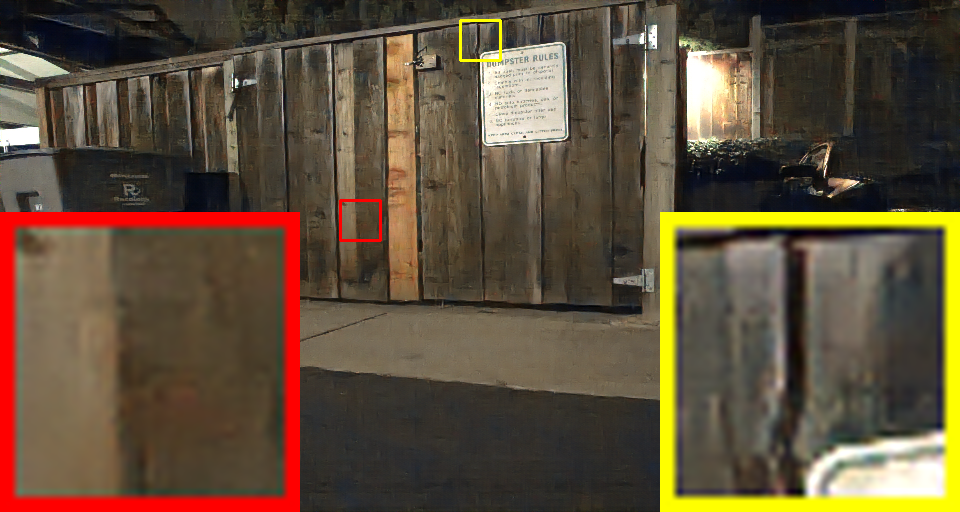}
		\vspace{-1.5em}
		\caption*{\small  \bf SNR+Ours}
	\end{subfigure}
	\begin{subfigure}[c]{0.11\textwidth}
		\centering
		\includegraphics[width=0.8in]{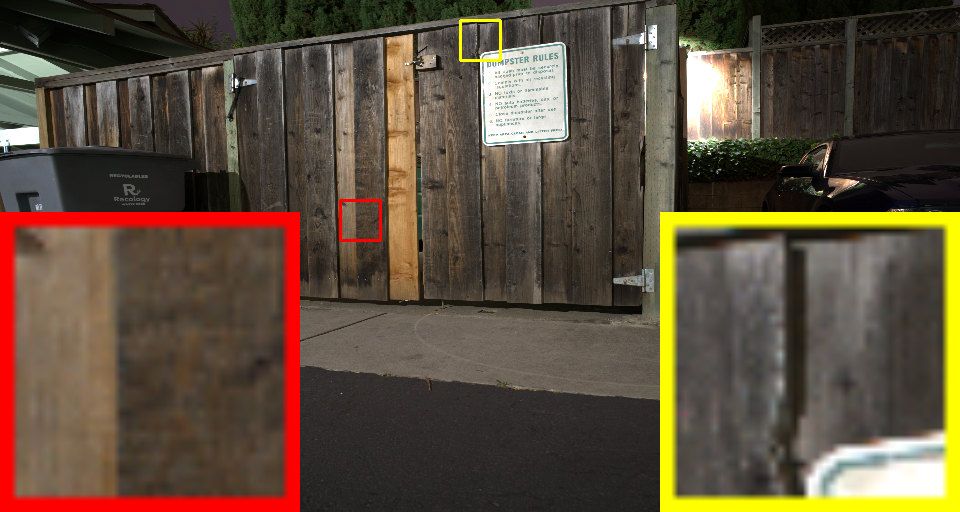}
		\vspace{-1.5em}
		\caption*{\small Ground-truth}
	\end{subfigure}  \\
 
\vspace{-0.1in}
	\caption{Comparisons on LOL-real (top) and SID (bottom). Results with ``Ours'' have less noise and clearer visibility.}
	\label{fig:cmp_llie}
\end{figure}

\begin{table}[t]
\centering
{\Huge
 \resizebox{1.0\linewidth}{!}{
		\begin{tabular}{c|cc|cc|cc|cc}
			\toprule[1pt]
		&\multicolumn{4}{c|}{LOL-real} & \multicolumn{4}{c}{SID} \\
  \cline{2-9}
  & \multicolumn{2}{c|}{URetinex} & \multicolumn{2}{c|}{SNR} &\multicolumn{2}{c|}{URetinex} & \multicolumn{2}{c}{SNR} \\
          &  PSNR & SSIM & PSNR & SSIM& PSNR & SSIM& PSNR & SSIM\\
            \hline
          w/o SP, with CA and SA  &23.45 & 0.868& 24.25& 0.886&21.98 & 0.619&23.02 &0.620 \\
          with SP, w/o CA, with SA &22.10 & 0.856& 24.05& 0.875& 22.05&\textbf{0.623} &22.93 &0.624 \\
          with SP and CA, w/o SA &23.76 & 0.850& 23.86&0.879 & 21.92&0.620 &23.07 &0.621 \\
          {Large $\mathcal{R}$ w/o SP/CA/SA} & 22.74&0.857 &24.51 &0.881 & 22.06& 0.621&23.04 &0.627 \\
          {w/o Position Embedding $\mathcal{S}$} &23.66 & 0.843&24.13 & 0.874&22.13 & 0.620&22.92 &0.622 \\
          {SNR Value as Mask} &22.66 & 0.855&24.77 &0.887 & 22.01& 0.617& 22.94&0.627 \\
          {Use 1D Priors via Con.} &23.01 &0.853 &23.83 &0.878 & 22.07&0.622 & 22.93&0.628 \\
          {Use 2D Priors via Con.} &22.68 & 0.862&24.11 &0.880 &22.08 & 0.618&23.06 & 0.625\\
          {Full Setting} &\textbf{24.70} & \textbf{0.878}& \textbf{25.50}&\textbf{0.892} & \textbf{22.34}& \textbf{0.623}& \textbf{23.34}&\textbf{0.630} \\
          \bottomrule[1pt]
	\end{tabular}}
\vspace{-0.1in}
	\caption{Ablation study results. We adopt CLIP as the pre-trained model. ``SP" denote PTG-SVE, ``CA" and ``SA" denote spatial- and channel attentions in PTG-CSA. Con. means Concatenation.}
\vspace{0.1in}
	\label{comparison4}
}

 \resizebox{1.0\linewidth}{!}{
		\begin{tabular}{c|p{1.4cm}<{\centering}p{1.4cm}<{\centering}p{1.4cm}<{\centering}|p{1.4cm}<{\centering}p{1.4cm}<{\centering}p{1.4cm}<{\centering}}
			\toprule[1pt]
                Datasets & \multicolumn{3}{c|}{LOL-real} & \multicolumn{3}{c}{SID}\\
                \hline
			Methods  & ZeroDCE & RUAS& SCI & ZeroDCE & RUAS & SCI\\
			PSNR   &18.06 &  18.37 &20.28 & 18.08 & 18.44& 19.09  \\
			SSIM &0.580 &  0.723 &0.752 & 0.576 &0.581 &0.585 \\
            \hline
            Methods  & +Ours-c & +Ours-c & +Ours-c & +Ours-c & +Ours-c & +Ours-c \\
		PSNR   &\textbf{18.79} &  \textbf{19.53} &\textbf{21.62}  & \textbf{18.65} & \textbf{18.93}& \textbf{19.61}  \\
            SSIM &\textbf{0.614} &  \textbf{0.747} &\textbf{0.781} & \textbf{0.593} &\textbf{0.590} &\textbf{0.598} \\
			\bottomrule[1pt]
	\end{tabular}}
  \vspace{-0.1in}
	\caption{Quantitative comparison on the LOL-real and SID dataset for unsupervised methods. We adopt CLIP as the pre-trained model here.}
  \vspace{-0.1in}
	\label{unsupervised}
\end{table}

\subsection{Low-light Image Enhancement}

\noindent\textbf{Comparison.}
We choose current SOTA low-light image enhancement methods as the baselines (UHD~\cite{wang2023ultra}, URetinex~\cite{wu2022uretinex}, SNR~\cite{xu2022snr}), and apply our refinement module for these baselines to see if their performance can be improved.
The priors are chosen from the CLIP~\cite{radford2021learning}, BLIP2~\cite{li2023blip}, Stable Diffusion~\cite{rombach2022high}, and pre-trained restoration models (trained on another dataset, as SDSD~\cite{wang2021seeing,xu2023deep}). We denote these results as $-c$, $-b$, $-s$, and $-r$, respectively.
In Table~\ref{comparison1}, we observe that combining these priors with our refinement module significantly improves the performance of the baselines.
Additionally, Fig.~\ref{fig:cmp_llie} provides visual comparisons.

Moreover, we conducted an experiment by adding the refinement module to the intermediate layer of $\mathcal{F}$, refining features of the target model. The refinement module is added to the deepest feature layer for efficiency, producing the residual feature map and the mask information for refinement. These results are denoted as $-f$. The improvement achieved by this operation is also evident as displayed in Table~\ref{comparison1}.

\vspace{1mm}
\noindent\textbf{Comparison with Other Priors.}
Some works, such as SKF~\cite{wu2023learning} and SMG~\cite{xu2023low}, utilize additional information like semantic maps, edge maps, and depth maps to enhance low-light image enhancement results. However, these methods require supervision with paired multi-modal information, whereas our method does not. Additionally, as shown in Table~\ref{comparison3}, our approach achieves better performance improvement for a given target model. Notably, the improvements achieved by other methods are based on large additional parameters, while our approach only uses a lightweight refinement module $<$ 1M.

\begin{table}
\begin{center}
\resizebox{1.0\linewidth}{!}{
\begin{tabular}{l| c c| c c| c c}
\toprule[0.15em]
 \textbf{Method} & PSNR~$\textcolor{black}{\uparrow}$ & SSIM~$\textcolor{black}{\uparrow}$ & PSNR~$\textcolor{black}{\uparrow}$ & SSIM~$\textcolor{black}{\uparrow}$ & PSNR~$\textcolor{black}{\uparrow}$ & SSIM~$\textcolor{black}{\uparrow}$ \\
\hline
  & \multicolumn{2}{c|}{\textbf{Test100}}&\multicolumn{2}{c|}{\textbf{Rain100H}}&\multicolumn{2}{c}{\textbf{Rain100L}}\\
SPAIR & {30.35} & {0.909} & {30.95} & {0.892} & {36.93} & {0.969}  \\
SPAIR+Ours-c  &\textbf{30.62}  &\textbf{0.917}  &\textbf{31.20} &\textbf{0.901} &\textbf{37.26} &\textbf{0.973} \\ \hline
Restormer& {32.00} & {0.923} & {31.46} & {0.904} & {38.99} & {0.978}  \\
Restormer+Ours-c  & \textbf{32.30} &\textbf{0.934}  &\textbf{31.77} &\textbf{0.913} &\textbf{39.27}&\textbf{0.985}  \\

\hline
& \multicolumn{2}{c|}{\textbf{Test2800}}&\multicolumn{2}{c|}{\textbf{Test1200}}&\multicolumn{2}{c}{\textbf{Average}}\\
SPAIR &33.34 & 0.936 & {33.04} &{0.922} & {32.91} & {0.926} \\
SPAIR+Ours-c  &\textbf{33.58}  &\textbf{0.942} &\textbf{33.35}  &\textbf{0.924}  & \textbf{33.16} &\textbf{0.932} \\ \hline
Restormer& {34.18} & {0.944} & {33.19} & {0.926} & {33.96} & {0.935} \\
Restormer+Ours-c  &\textbf{34.47}  & \textbf{0.951}&\textbf{33.48}  & \textbf{0.929} & \textbf{34.24} &\textbf{0.943} \\
\bottomrule[0.1em]
\end{tabular}}
\vspace{-0.1in}
\caption{Image deraining results. 
}
\label{table:deraining}
\vspace{-0.1in}
\end{center}
\end{table}

\begin{figure}[t]
\centering
	\begin{subfigure}[c]{0.23\textwidth}
		\centering
		\includegraphics[width=1.5in, height=0.8in]{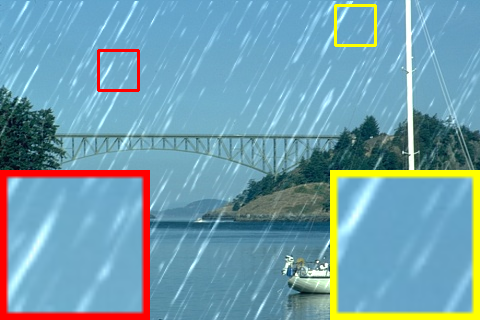}
		\caption*{\small Input}
	\end{subfigure}
	\begin{subfigure}[c]{0.23\textwidth}
		\centering
		\includegraphics[width=1.5in, height=0.8in]{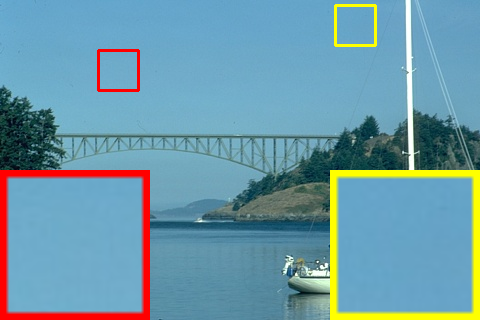}
		\caption*{\small Ground-truth}
	\end{subfigure}  
	\begin{subfigure}[c]{0.23\textwidth}
		\centering
		\includegraphics[width=1.5in, height=0.8in]{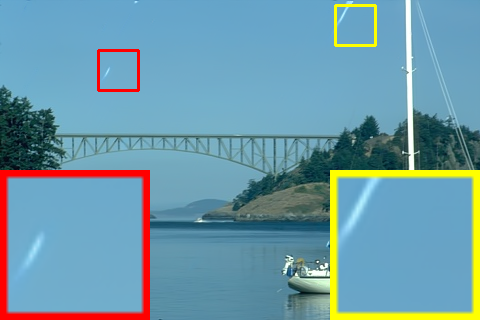}
		\caption*{\small Restormer}
	\end{subfigure}
	\begin{subfigure}[c]{0.23\textwidth}
		\centering
		\includegraphics[width=1.5in, height=0.8in]{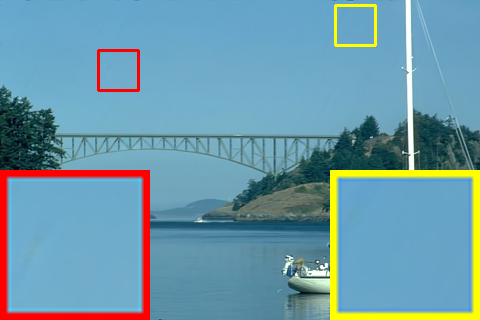}
		\caption*{\small \bf Restormer+Ours}
	\end{subfigure}
	\vspace{0.2em} \\
 
	\vspace{-0.1in}
	\caption{Visual comparison on Rain100H showing the effects of our strategy. 
 }
	\vspace{-0.1in}
	\label{fig:derain}
\end{figure}

\noindent\textbf{Ablation Study: Ablation of Different Components.}
We first set experiments by deleting different components from our framework, including PTG-SVE (abbreviated as ``SP"), and spatial-channel attentions with priors that are abbreviated as ``CA" and ``SA", respectively.
As shown in Table~\ref{comparison4}, deleting any component will lead to a performance drop.

We conduct experiments without SP, CA, or SA to analyze whether additional parameters or priors take a prominent role. The short-range and long-range results are fused via a simple sum, and the spatial-channel attention is conducted using only the features themselves. Additionally, we increase the feature channel number fourfold to add more parameters. The results, denoted as ``Large $\mathcal{R}$ w/o SP/CA/SA" in Table~\ref{comparison4}, are still lower than our full setting, indicating the effectiveness of our proposed approach over simply increasing parameters.

In addition, we perform an experiment by removing the learnable position embeddings $\mathcal{S}_m$ and $\mathcal{S}_c$, denoted as ``w/o Position Embedding for Priors" in Table~\ref{comparison4}. This comparison highlights the importance of using spatial-aware representations for the pre-trained features.

\vspace{1mm}
\noindent\textbf{Ablation Study: SNR Value as Mask.}
In comparison to previous methods that directly use the SNR value as the mask to fuse the short- and long-range results, our approach utilizes pre-trained priors to automatically discover restoration-related information and formulate the fusion mask adaptively. In this ablation study, we demonstrate that our strategy outperforms the direct SNR-based approach, as shown in Table~\ref{comparison4}.

\begin{table}[!t]
\begin{center}
\label{table:deblurring}
\resizebox{1.0\linewidth}{!}{
\begin{tabular}{l| cc | cc | cc | cc }
\toprule[0.15em]
 \multirow{2}{1cm}{\textbf{Method}} & \multicolumn{2}{c|}{\textbf{GoPro}} & \multicolumn{2}{c|}{\textbf{HIDE}} & \multicolumn{2}{c|}{\textbf{RealBlur-R}} & \multicolumn{2}{c}{\textbf{\textbf{RealBlur-J}}} \\
  & PSNR & {SSIM} & PSNR & {SSIM} & PSNR & {SSIM} & PSNR &{SSIM}\\
\hline
MPRNet & {32.66} & {{0.959}} &	{30.96} &{{0.939}} & {35.99} &  {{0.952}} & 28.70& {0.873} \\
MPRNet+Ours-c & \textbf{32.87}&\textbf{0.964} &\textbf{31.19}  &\textbf{0.943} &\textbf{36.25}  &\textbf{0.960} &\textbf{28.98} &\textbf{0.881}\\ \hline
Restormer & {32.92} & {{0.961}} & {31.22} & {{0.942}} & {36.19} & {{0.957}} & {28.96} & {{0.879}}\\
Restormer+Ours-c & \textbf{33.18}& \textbf{0.966}&\textbf{31.51}  & \textbf{0.950}&\textbf{36.47}  &\textbf{0.962} & \textbf{29.21}&\textbf{0.883}\\
\bottomrule[0.1em]
\end{tabular}}
\vspace{-0.1in}
\caption{Single-image motion deblurring results. 
}
\vspace{-0.1in}
\label{tab:motion}
\end{center}
\end{table}

\begin{figure}[t]
	\centering

 \begin{subfigure}[c]{0.23\textwidth}
		\centering
		\includegraphics[width=1.5in, height=0.8in]{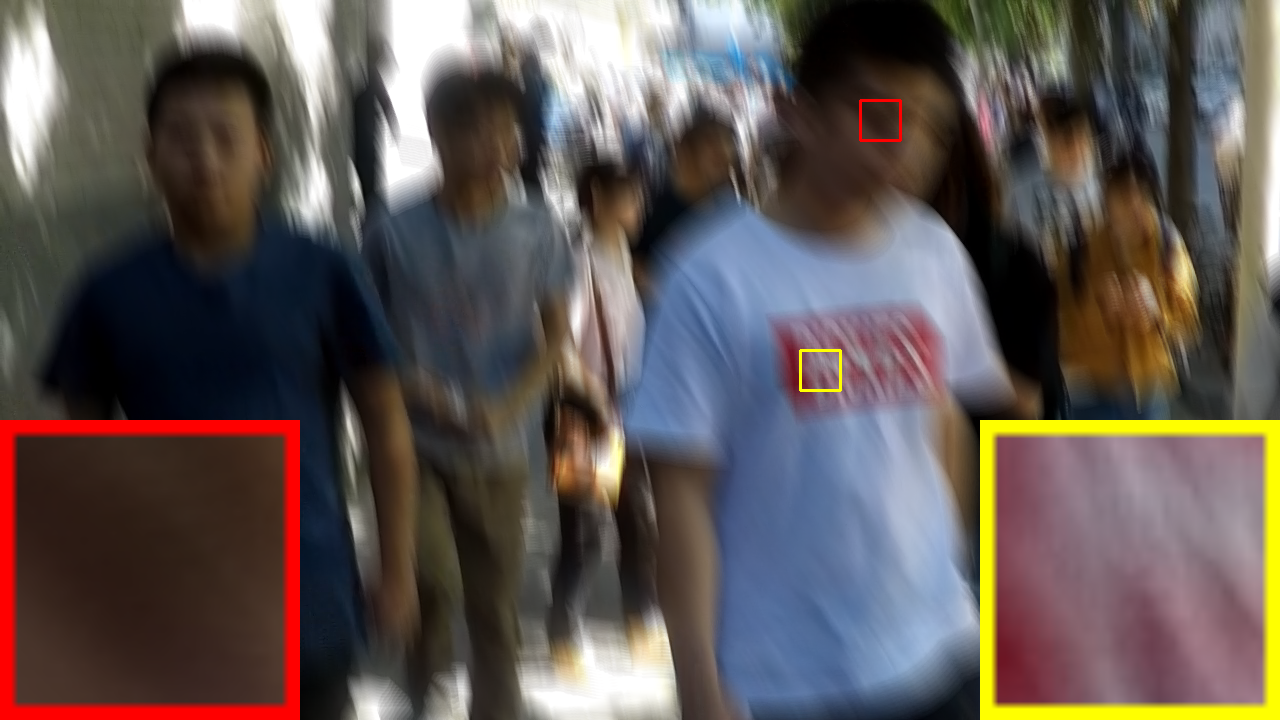}
		\caption*{\small Input}
	\end{subfigure}
	\begin{subfigure}[c]{0.23\textwidth}
		\centering
		\includegraphics[width=1.5in, height=0.8in]{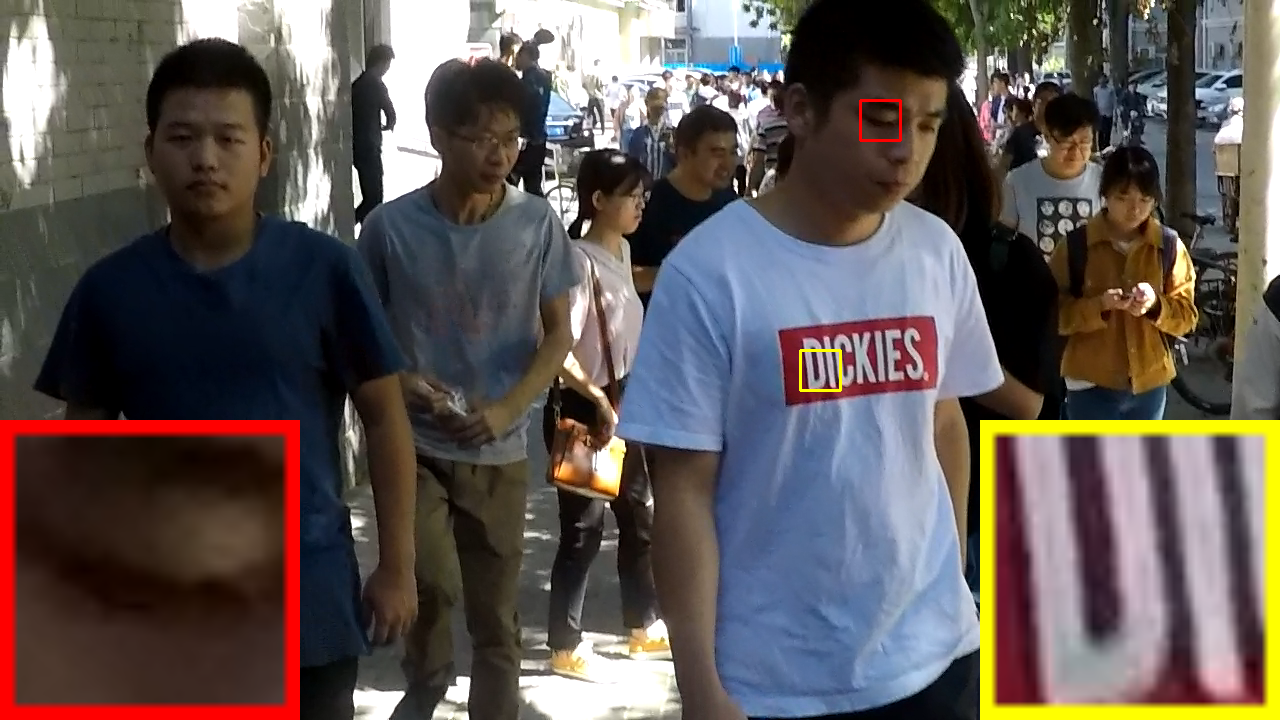}
		\caption*{\small Ground-truth}
	\end{subfigure} 
	\begin{subfigure}[c]{0.23\textwidth}
		\centering
		\includegraphics[width=1.5in, height=0.8in]{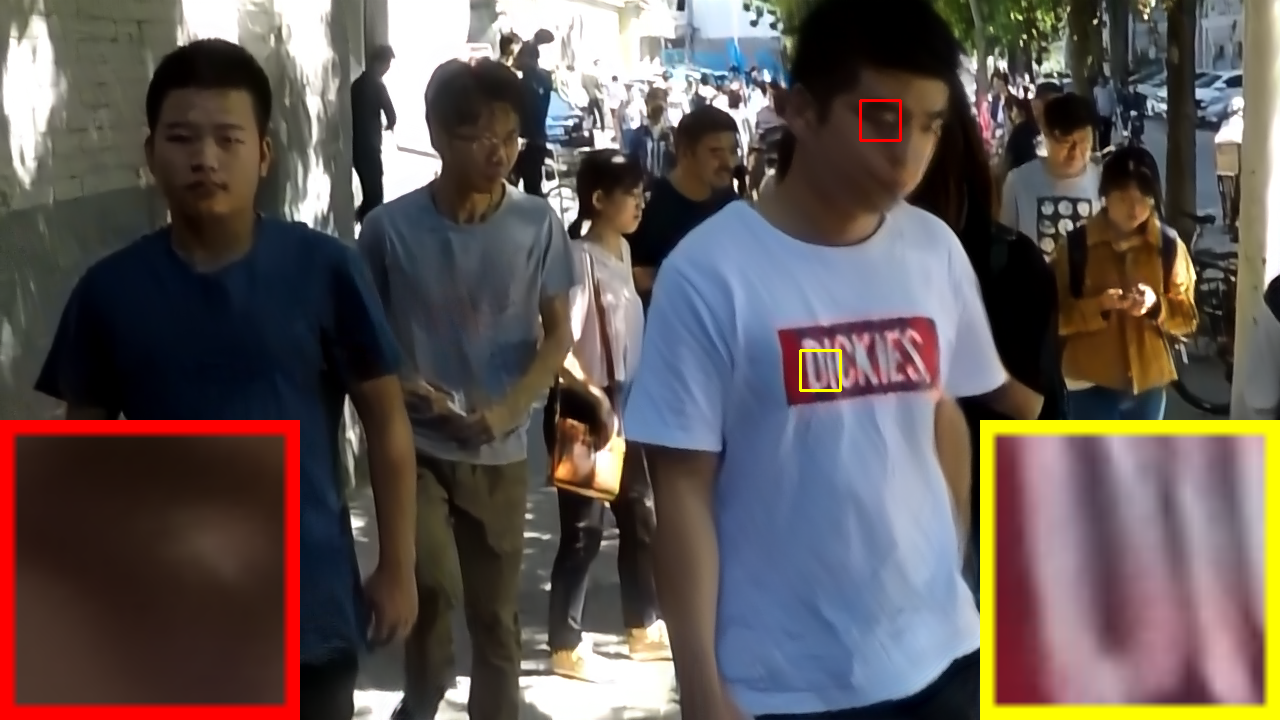}
		\caption*{\small Restormer}
	\end{subfigure}
	\begin{subfigure}[c]{0.23\textwidth}
		\centering
		\includegraphics[width=1.5in, height=0.8in]{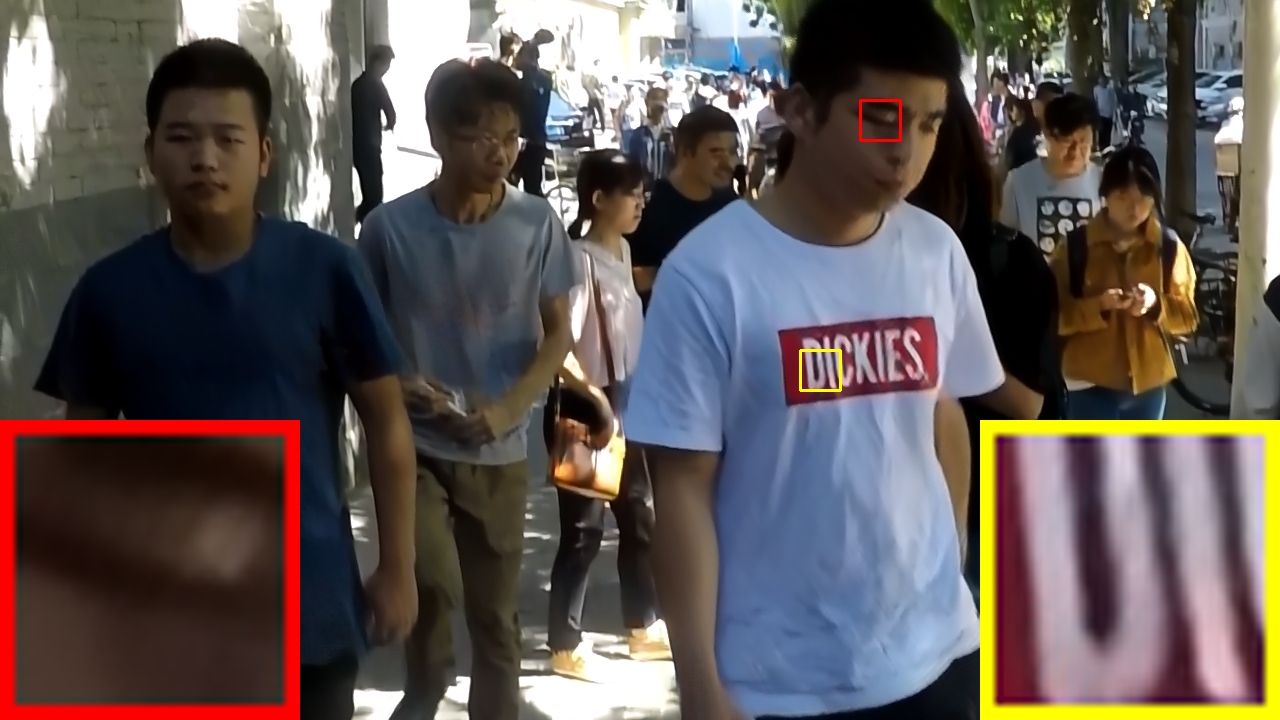}
		\caption*{\small \bf Restormer+Ours}
	\end{subfigure}
	\vspace{-0.1in}
	\caption{Visual comparison on HIDE. 
 }
	\vspace{0.1in}
	\label{fig:motion}

	\centering
	\begin{subfigure}[c]{0.23\textwidth}
		\centering
		\includegraphics[width=1.5in, height=0.8in]{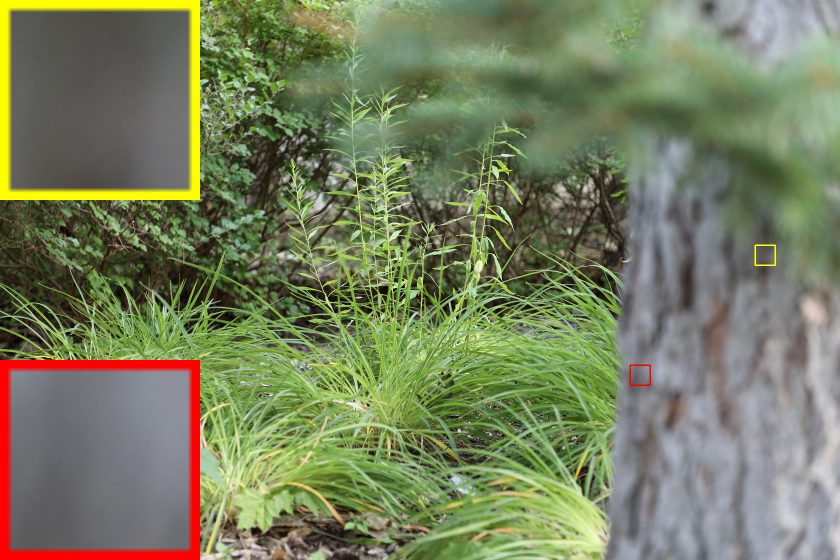}
		\caption*{\small Input}
	\end{subfigure}
	\begin{subfigure}[c]{0.23\textwidth}
		\centering
		\includegraphics[width=1.5in, height=0.8in]{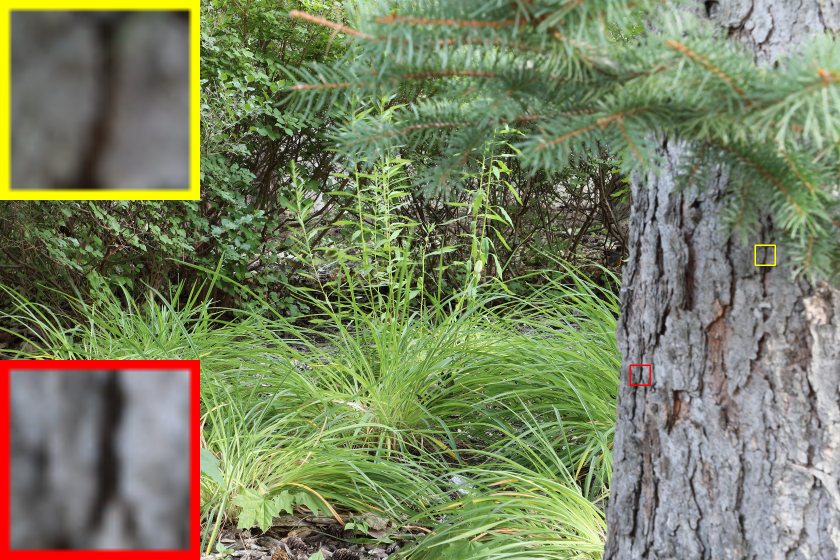}
		\caption*{\small Ground-truth}
	\end{subfigure} 
	\begin{subfigure}[c]{0.23\textwidth}
		\centering
		\includegraphics[width=1.5in, height=0.8in]{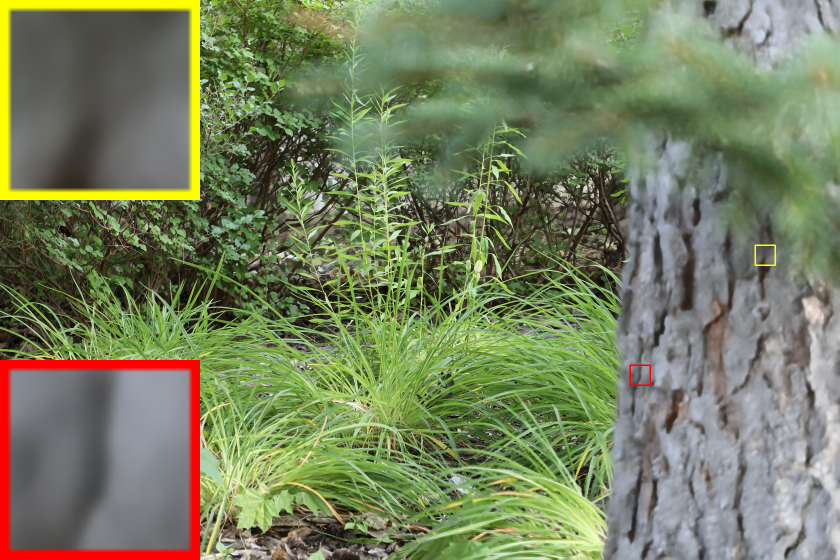}
		\caption*{\small GRL}
	\end{subfigure}
	\begin{subfigure}[c]{0.23\textwidth}
		\centering
		\includegraphics[width=1.5in, height=0.8in]{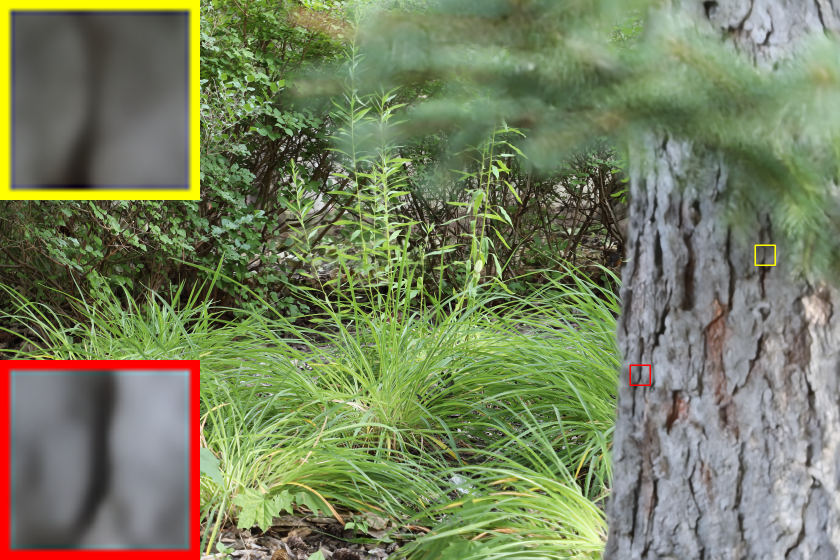}
		\caption*{\small \bf GRL+Ours}
	\end{subfigure} \\

	\vspace{-0.1in}
	\caption{Visual comparison on single-image defocus deblurring.
 }
	\vspace{-0.2in}
	\label{fig:defocus}
\end{figure}

\vspace{1mm}
\noindent\textbf{Ablation Study: Alternatives of Using Priors.}
In this study, we demonstrate the difficulty of directly aligning priors to the restoration features. We conduct an experiment where the priors are concatenated with the features in the refinement module to implement different components. However, the improvement obtained with this direct approach is not as significant as our proposed method, as shown in Table~\ref{comparison4}. This is because the different features are heterogeneous with the restoration features, even when the priors are adopted as 2D feature maps. This study highlights the importance of our novel strategy of employing these priors.

\vspace{1mm}
\noindent\textbf{$\mathcal{R}$ for Unsupervised Approach.}
Different from existing refinement methods that need supervision for learning the additional features (e.g., SKF needs the semantic ground truth of the normal-light data, SMG needs the depth and edge information of the normal-light data), our approach does not require the feature of the normal-light data during both training and inference.
We only need the feature that is extracted from $I_d$ with the pre-trained model $\mathcal{G}$ during the training.
Also, the loss function for training the refinement module can be set the same as that of the target model.
Thus, the unsupervised training of the target model can also be adopted in our framework.
As shown in Table~\ref{unsupervised}, our method can successfully improve the performance of various unsupervised low-light image enhancement methods with different unsupervised loss terms, including EnGAN~\cite{jiang2021enlightengan}, ZeroDCE~\cite{Zero-DCE}, RUAS~\cite{liu2021retinex}, and SCI~\cite{ma2022toward}.

\subsection{Other Restoration Tasks}
In this section, we conduct experiments using CLIP as the pre-trained model ($-c$). CLIP is chosen for its efficiency and convenience compared to other pre-trained models. 

\vspace{1mm}
\noindent\textbf{Deraining.}
For deraining tasks, we use SOTA methods such as SPAIR~\cite{purohit2021spatially_spair} and Restormer~\cite{zamir2022restormer} as baselines.
We compute PSNR/SSIM values using the Y channel in the YCbCr color space, similar to existing methods. Table~\ref{table:deraining} demonstrates that our approach improves the performance of these existing methods and consistently achieves significant performance gains across all five datasets. The qualitative comparison results are shown in Fig.~\ref{fig:derain}.

\vspace{1mm}
\noindent\textbf{Motion Deblurring.}
We analyze our approach for deblurring tasks on synthetic datasets (GoPro, HIDE) and real-world datasets (RealBlur-R, RealBlur-J). 
The baselines include MPRNet~\cite{Zamir_2021_CVPR_mprnet} and Restormer~\cite{zamir2022restormer}.
Table~\ref{tab:motion} demonstrates that our approach improves the performance of all these methods on all four benchmark datasets.
Although the enhanced network is trained only on the GoPro dataset, it shows more robust generalization to other datasets. Qualitative comparisons are shown in Fig.~\ref{fig:motion}, further supporting our claim.

\vspace{1mm}
\noindent\textbf{Defocus Deblurring.}
Table~\ref{table:dpdeblurring} presents the image fidelity scores of SOTA approaches on the DPDD dataset~\cite{abdullah2020dpdd}, including IFAN~\cite{Lee_2021_CVPRifan}, Restormer~\cite{zamir2022restormer}, and GRL~\cite{li2023efficient}. Our refinement module achieves significant performance improvement for these SOTA schemes in both single-image and dual-pixel defocus deblurring settings across all scene categories. 
The qualitative results are depicted in Fig.~\ref{fig:defocus}.

\begin{table}[!t]
\begin{center}
\Huge
\resizebox{1.0\linewidth}{!}{
\begin{tabular}{l | c c c  | c c c  | c c c }
\toprule[0.15em]
   \multirow{2}{1cm}{\textbf{Method}} & \multicolumn{3}{c|}{\textbf{Indoor Scenes}} & \multicolumn{3}{c|}{\textbf{Outdoor Scenes}} & \multicolumn{3}{c}{\textbf{Combined}} \\
\cline{2-10}
    & PSNR & SSIM& LPIPS  & PSNR & SSIM& LPIPS  & PSNR & SSIM~& LPIPS   \\
\midrule[0.15em]
IFAN$_S$ & {28.11}  & {0.861}   & {0.179}  & {22.76}  & {0.720}   & {0.254}  & {25.37} & {0.789} & {0.217}\\
IFAN$_S$+Ours-c &\textbf{28.32} &\textbf{0.870}  &\textbf{0.171} &\textbf{23.08} &\textbf{0.727}  &\textbf{0.248} &\textbf{25.72} &\textbf{0.795}  &\textbf{0.213}\\
{Restormer}$_S$& {28.87}  & {0.882}  & {0.145} & {23.24}  & {0.743}  & {0.209}  & {25.98}  & {0.811}   & {0.178}   \\
{Restormer}$_S$+Ours-c & \textbf{29.17}&\textbf{0.890}  &\textbf{0.141} &\textbf{23.43} &\textbf{0.749}  &\textbf{0.206} &\textbf{26.13} &\textbf{0.816}  &\textbf{0.165}\\
GRL$_S$-B &29.06 & 0.886  &0.139 & 23.45 &0.761 & 0.196& 26.18& 0.822& 0.168\\
GRL$_S$-B +Ours-c &\textbf{29.30} &\textbf{0.894}  &\textbf{0.133} &\textbf{23.67} &\textbf{0.768} &\textbf{0.189} &\textbf{26.45} &\textbf{0.828}  &\textbf{0.161}\\
\midrule[0.1em]

IFAN$_D$ & {28.66} & {0.868}  & {0.172} & {23.46} & {0.743} & {0.240} & {25.99} & {0.804}  & {0.207} \\
IFAN$_D$+Ours-c& \textbf{28.94} &\textbf{0.875}  &\textbf{0.167} &\textbf{23.70} &\textbf{0.748} &\textbf{0.235} &\textbf{26.20} &\textbf{0.811}  &\textbf{0.203}\\
{Restormer}$_D$& {29.48}  & {0.895}   & {0.134} & {23.97}  & {0.773}   & {0.175}  & {26.66}  & {0.833}   & {0.155} \\
{Restormer}$_D$+Ours-c&\textbf{29.79} &\textbf{0.902} &\textbf{0.131} &\textbf{24.23} &\textbf{0.778}  &\textbf{0.155} & \textbf{26.89}&\textbf{0.840} &\textbf{0.153}\\
GRL$_D$-B&29.83& 0.903& 0.114& 24.39& 0.795& 0.150& 27.04& 0.847& 0.133\\
GRL$_D$-B+Ours-c &\textbf{29.96} &\textbf{0.911}  &\textbf{0.110} &\textbf{24.62} &\textbf{0.803}  &\textbf{0.145} & \textbf{27.27}& \textbf{0.855} &\textbf{0.128}\\

\bottomrule[0.1em]
\end{tabular}}
\vspace{-0.1in}
\caption{Defocus deblurring comparisons on the DPDD testset (containing 37 indoor and 39 outdoor scenes). \textbf{S:} single-image defocus deblurring. \textbf{D:} dual-pixel defocus deblurring. 
}
\label{table:dpdeblurring}
\end{center}

\resizebox{1.0\linewidth}{!}{
\begin{tabular}{l | c c c | c c c | c c c}
\toprule[0.15em]
   \multirow{2}{1cm}{\textbf{Method}}& \multicolumn{3}{c|}{\textbf{Set12}} & \multicolumn{3}{c|}{\textbf{BSD68}} & \multicolumn{3}{c}{\textbf{Urban100}} \\
 \cline{2-10}
   & $\sigma$$=$$15$ & $\sigma$$=$$25$ & $\sigma$$=$$50$ & $\sigma$$=$$15$ & $\sigma$$=$$25$ & $\sigma$$=$$50$ & $\sigma$$=$$15$ & $\sigma$$=$$25$ & $\sigma$$=$$50$ \\
\midrule[0.15em]
DRUNet  & {33.25} & {30.94} & {27.90} & {31.91} & {29.48} & {26.59} & {33.44} & {31.11} & {27.96}\\ %top
DRUNet+Ours-c &\textbf{33.51} &\textbf{31.18} &\textbf{28.27} &\textbf{32.20} &\textbf{29.73} &\textbf{26.84} &\textbf{33.65} & \textbf{31.34}&\textbf{28.16} \\
Restormer & {33.35}	& {31.04}	& {28.01} & {31.95}	& {29.51}	& {26.62} & {33.67}	& {31.39}	& {28.33}\\ 
Restormer+Ours-c & \textbf{33.57}& \textbf{31.28}&\textbf{28.36} &\textbf{32.11} &\textbf{29.78} & \textbf{26.91}& \textbf{33.96}&\textbf{31.67} &\textbf{28.58} \\
\midrule[0.1em]
 Restormer& {33.42} & {31.08} & {28.00} & {31.96} & {29.52}& {26.62}&{33.79} & {31.46}& {28.29}\\ 
 Restormer+Ours-c &\textbf{33.70} &\textbf{31.29} &\textbf{28.35} & \textbf{32.24}&\textbf{29.81} & \textbf{26.86}&\textbf{33.97} &\textbf{31.73} &\textbf{28.58} \\
GRL-B & 33.47 & 31.12 & 28.03 & 32.00 & 29.54 & 26.60 &34.09& 31.80& 28.59\\
GRL-B+Ours-c&\textbf{33.74} &\textbf{31.30} &\textbf{28.37} &\textbf{32.29} & \textbf{29.76}&\textbf{26.91} &\textbf{34.22} &\textbf{31.95} &\textbf{28.74} \\
\bottomrule[0.1em]
\end{tabular}}
\vspace{-0.1in}
\caption{Gaussian grayscale image denoising comparisons. Top super rows: learning a single model to handle various noise levels. Bottom super rows: training a separate model for each noise level. 
} 
\label{table:gauss-gray}
\vspace{-0.1in}
\end{table}

\begin{figure}[t]
	\centering
	\begin{subfigure}[c]{0.23\textwidth}
		\centering
		\includegraphics[width=1.5in, height=0.8in]{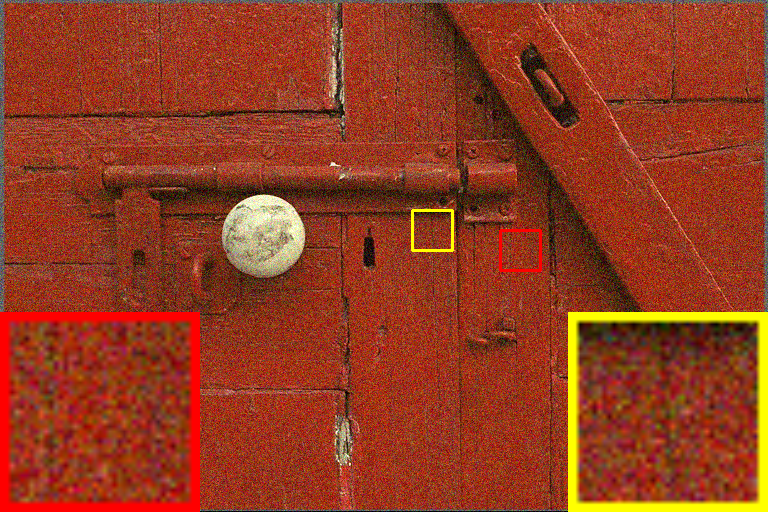}
		%\vspace{-2mm}
		\caption*{\small Input}
	\end{subfigure}
	\begin{subfigure}[c]{0.23\textwidth}
		\centering
		\includegraphics[width=1.5in, height=0.8in]{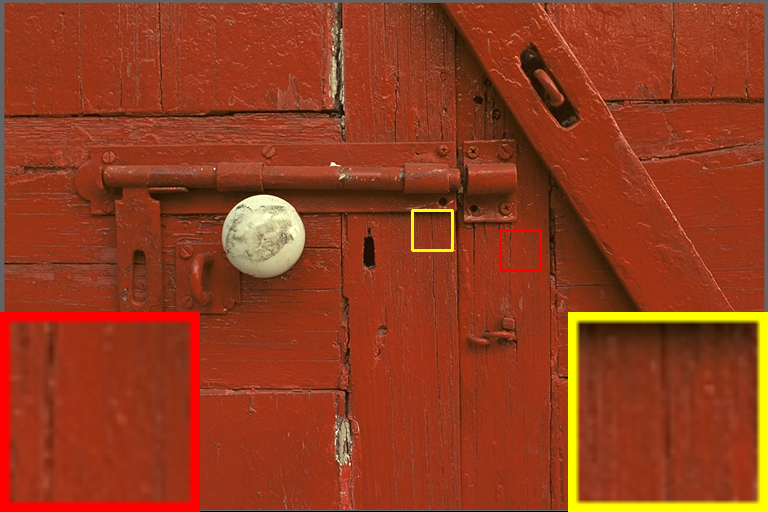}
		%\vspace{-2mm}
		\caption*{\small Ground-truth}
	\end{subfigure}  
	\begin{subfigure}[c]{0.23\textwidth}
		\centering
		\includegraphics[width=1.5in, height=0.8in]{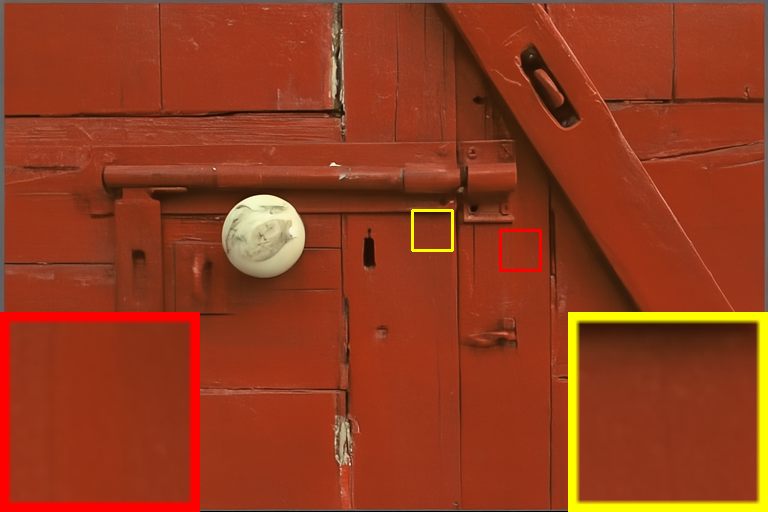}
		%\vspace{-2mm}
		\caption*{\small GRL}
	\end{subfigure}
	\begin{subfigure}[c]{0.23\textwidth}
		\centering
		\includegraphics[width=1.5in, height=0.8in]{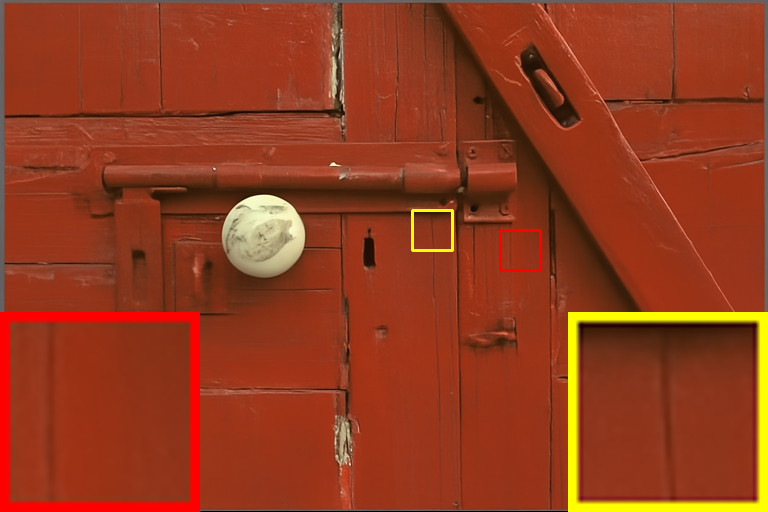}
		%\vspace{-2mm}
		\caption*{\small  \bf GRL+Ours}
	\end{subfigure}
	\vspace{0.2em} \\
	
	\begin{subfigure}[c]{0.11\textwidth}
		\centering
		\includegraphics[width=0.8in]{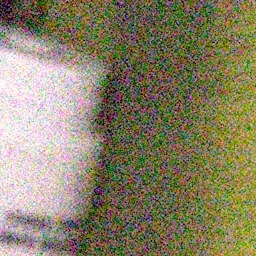}
		\vspace{-1.5em}
		\caption*{\small Input}
	\end{subfigure}
	\begin{subfigure}[c]{0.11\textwidth}
		\centering
		\includegraphics[width=0.8in]{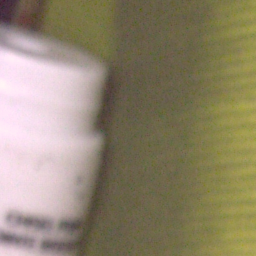}
		\vspace{-1.5em}
		\caption*{\small Ground-truth}
	\end{subfigure}  
	\begin{subfigure}[c]{0.11\textwidth}
		\centering
		\includegraphics[width=0.8in]{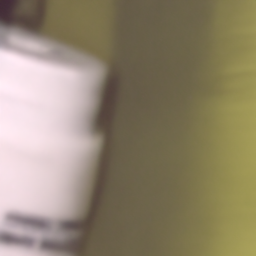}
		\vspace{-1.5em}
		\caption*{\small Restormer}
	\end{subfigure}
	\begin{subfigure}[c]{0.11\textwidth}
		\centering
		\includegraphics[width=0.8in]{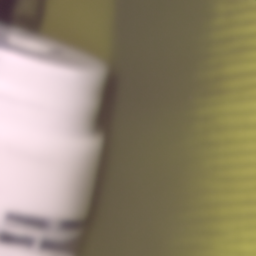}
		\vspace{-1.5em}
		\caption*{\small  \bf +Ours}
	\end{subfigure}\\
	
	\vspace{-0.1in}
	\caption{Visual comparisons on Kodak (top) and SIDD (bottom). 
	}
	\label{fig:denoise}
	
	\begin{center}
		\includegraphics[width=1.0\linewidth]{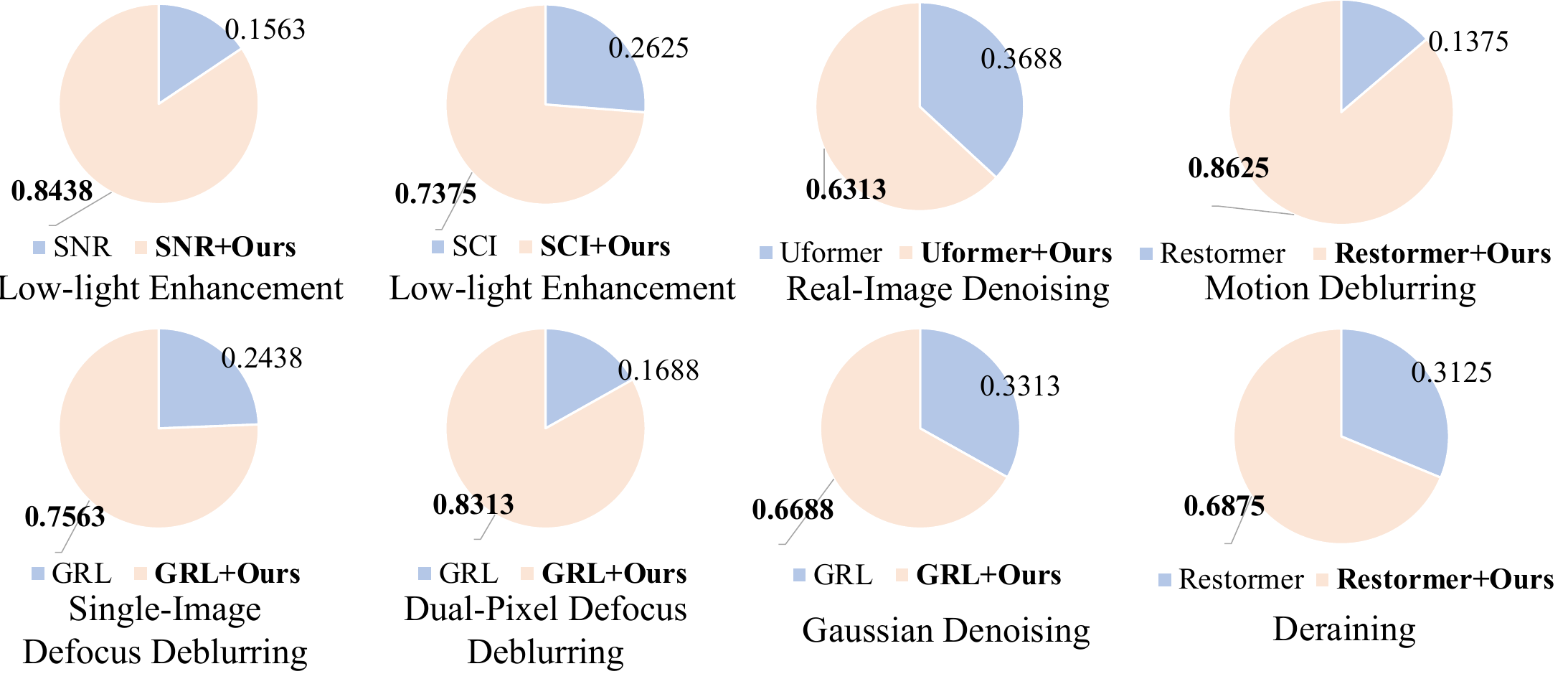}
	\end{center}
	\vspace{-0.25in}
	\caption{The user study results show that our strategy can effectively improve the performance of restoration approaches in terms of human subjective evaluation.
	}
	\vspace{-0.2in}
	\label{fig:user_study}
\end{figure}

\vspace{1mm}
\noindent\textbf{Gaussian Denoising.}
We conduct denoising experiments on synthetic benchmark datasets with additive white Gaussian noise. We choose DRUNet~\cite{zhang2021DPIR}, Restormer~\cite{zamir2022restormer}, and GRL~\cite{li2023efficient} as baselines, which are SOTA approaches in denoising. Tables~\ref{table:gauss-gray} and \ref{table:gauss-color} present PSNR scores of different approaches on grayscale and color image denoising, respectively, for noise levels of 15, 25, and 50. We evaluate two experimental settings: (1) learning a single model to handle various noise levels and (2) learning separate models for each noise level. Our method achieves significant performance enhancement for all these methods under both experimental settings on different datasets and noise levels. The visual results are shown in Fig.~\ref{fig:denoise}, showing the effectiveness of our strategy.

\begin{table}[t]
\centering
\Huge
\label{table:colordenoising}
\resizebox{1.0\linewidth}{!}{
\begin{tabular}{l | c c c | c c c | c c c | c c c}
\toprule[0.15em]
   \multirow{2}{1cm}{\textbf{Method}}& \multicolumn{3}{c|}{\textbf{CBSD68}} & \multicolumn{3}{c|}{\textbf{Kodak24}} & \multicolumn{3}{c|}{\textbf{McMaster}} & \multicolumn{3}{c}{\textbf{Urban100}} \\
 \cline{2-13}
    & $\sigma$$=$$15$ & $\sigma$$=$$25$ & $\sigma$$=$$50$ & $\sigma$$=$$15$ & $\sigma$$=$$25$ & $\sigma$$=$$50$ & $\sigma$$=$$15$ & $\sigma$$=$$25$ & $\sigma$$=$$50$ & $\sigma$$=$$15$ & $\sigma$$=$$25$ & $\sigma$$=$$50$ \\
\midrule[0.15em]
DRUNet  & {34.30} & {31.69} & {28.51} & {35.31} & {32.89} & {29.86} & {35.40} & {33.14} & {30.08} & {34.81} & {32.60} & {29.61} \\
+Ours-c &\textbf{34.54} &\textbf{31.97} &\textbf{28.76} &\textbf{35.58} &\textbf{33.15} &\textbf{29.97} &\textbf{35.71} &\textbf{33.50} &\textbf{30.25} &\textbf{35.10} &\textbf{32.82} &\textbf{29.83} \\
Restormer & {34.39} & {31.78} & {28.59} & {35.44} & {33.02} & {30.00} & {35.55} & {33.31} & {30.29} & {35.06} & {32.91} & {30.02}\\ 
+Ours-c &\textbf{34.63} &\textbf{32.04} &\textbf{28.88} &\textbf{35.65} &\textbf{33.26} & \textbf{30.15}&\textbf{35.86} &\textbf{33.64} & \textbf{30.63}& \textbf{35.26}&\textbf{33.22} &\textbf{30.21} \\
\midrule[0.1em]
Restormer & {34.40} & {31.79}& {28.60}& {35.47} & {33.04}& {30.01}& {35.61}& {33.34}& {30.30} & {35.13}& {32.96}& {30.02}\\ 
+Ours-c &\textbf{34.76} &\textbf{32.05} &\textbf{28.94} & \textbf{35.72}&\textbf{33.27} &\textbf{30.21} & \textbf{35.80}& \textbf{33.63}&\textbf{30.55} &\textbf{35.32} &\textbf{33.14} &\textbf{30.27} \\
GRL-B & 34.45& 31.82& 28.62& 35.43& 33.02& 29.93& 35.73& 33.46& 30.36& 35.54& 33.35& 30.46\\
+Ours-c &\textbf{34.73} &\textbf{32.07} &\textbf{28.90} &\textbf{35.71} &\textbf{33.24} &\textbf{30.18} &\textbf{35.96} &\textbf{33.75} &\textbf{30.62} &\textbf{35.70} &\textbf{33.57} &\textbf{30.64} \\
\bottomrule[0.1em]
\end{tabular}}
\vspace{-0.1in}
\caption{Gaussian color image denoising. Equivalent notation meanings (top and bottom rows) as those in Table~\ref{table:gauss-gray}.}
\label{table:gauss-color}
\vspace{-0.2in}

{
\Large
\begin{center}
\resizebox{1.0\linewidth}{!}{
\begin{tabular}{c| c| p{1.8cm}<{\centering}p{1.8cm}<{\centering}p{1.8cm}<{\centering}p{1.8cm}<{\centering}p{1.8cm}<{\centering}p{1.8cm}<{\centering} }
\toprule[0.15em]
\textbf{Dataset}& \textbf{ Method} & MPRNet & MPRNet + Ours-c & 
Uformer&Uformer + Ours-c &Restormer &Restormer + Ours-c \\
\midrule[0.15em]
\textbf{SIDD} & PSNR~$\textcolor{black}{\uparrow}$  & {\Large 39.71} &{\Large \textbf{39.93}} & {\Large 39.77}& {\Large \textbf{39.94}} & {\Large 40.02}&{\Large \textbf{40.22}} \\
  & SSIM~$\textcolor{black}{\uparrow}$ & {\Large 0.958} & {\Large \textbf{0.961}} & {\Large 0.959} &{\Large \textbf{0.962}} & {\Large 0.960} &{\Large \textbf{0.965}} \\
\bottomrule
\end{tabular}}
\vspace{-0.1in}
\caption{Real image denoising on the SIDD dataset.}
\label{table:realdenoising}
\vspace{-0.3in}
\end{center}
}
\end{table}

\vspace{1mm}
\noindent\textbf{Real Denoising.}
We also conduct denoising experiments on the real-world SIDD dataset, with MPRNet~\cite{Zamir_2021_CVPR_mprnet}, Uformer~\cite{wang2021uformer}, and Restormer~\cite{zamir2022restormer} as baselines. Table~\ref{table:realdenoising} demonstrates that our refinement method improves both PSNR and SSIM metrics. Notably, on the SIDD dataset, our refinement enables the SOTA approach Restormer to achieve a PSNR surpassing 40.2 dB. The visual comparison is shown in Fig.~\ref{fig:denoise}.

\vspace{1mm}
\noindent\textbf{User Study.}
Furthermore, we conducted a large-scale user study with an A/B test strategy involving 80 participants.
Each participant is asked to simultaneously see two restored results, i.e., baseline and baseline+ours, and gauge which one is better. As shown in Fig.~\ref{fig:user_study}, the results combined with our strategy are more preferred by the participants.

\section{Conclusion}
In this work, we explore the utilization of features from a pre-trained model to enhance the performance of a restoration model. By unifying the shapes of the pre-trained features, we introduce a novel refinement module PTG-RM that employs PTG-SVE and PTG-CSA mechanisms. Unlike existing strategies, we focus on formulating optimal operation ranges and attention strategies guided by the pre-trained features. The extensive experiments conducted on various tasks, datasets, and networks demonstrate the effectiveness and generalization ability of our approach. We believe that our proposed principle of discovering hidden useful information in pre-trained models can be applicable to other domains as well.

\noindent\textbf{Limitation and Future Work.}
While our proposed strategy has exhibited significant effects in enhancing the performance of diverse restoration networks across various architectures with its lightweight module, the extent of improvement appears to vary across different experiments. Some instances showcase noticeable enhancement, while others do not. Such differences correlate with the capacity of the target network and the difficulty/complexity of the target task. In future endeavors, we intend to delve into more effective approaches that specifically aid target restoration tasks. We aim to employ a tailored distillation framework to derive refined restoration feature priors, ultimately making significant strides beyond existing upper boundaries. We also aim to develop corresponding technical products.

\noindent
\textbf{Acknowledgements.} 
This work is supported by the Natural Science Foundation of Zhejiang Pvovince, China, under No. LD24F020002.
SK is partially supported by University of Macau (SRG2023-00044-FST).

{
    \small
    \bibliographystyle{ieeenat_fullname}
    \bibliography{main}
}


\end{document}